\documentclass{article}

\PassOptionsToPackage{numbers, compress}{natbib}

\usepackage[preprint]{neurips_data_2024}

\usepackage[utf8]{inputenc} 
\usepackage[T1]{fontenc}    
\usepackage{hyperref}       
\usepackage{url}            
\usepackage{booktabs}       
\usepackage{amsfonts}       
\usepackage{nicefrac}       
\usepackage{microtype}      
\usepackage{xcolor}         
\usepackage{graphicx}
\usepackage{hyperref}       
\usepackage{url}            
\usepackage{booktabs}       
\usepackage{amsfonts}       
\usepackage{nicefrac}       
\usepackage{microtype}      
\usepackage{bm}
\usepackage{amsmath}
\usepackage{amssymb}
\usepackage{mathtools}
\usepackage{amsthm}
\usepackage{multirow}
\usepackage{hyperref}
\usepackage{wrapfig}
\usepackage{multirow} 
\usepackage{array}    
\usepackage{colortbl}
\usepackage{wrapfig}  

\usepackage[linesnumbered,ruled,vlined]{algorithm2e}
\usepackage[ruled]{algorithm2e}
\usepackage{algpseudocode}  
\usepackage{amsmath} 
\usepackage{anyfontsize}
\usepackage{caption}
\usepackage{subfigure}
\usepackage{graphicx}
\usepackage{svg}
\usepackage{makecell}
\usepackage{listings}
\usepackage{hyperref}
\usepackage{authblk}

\title{Empowering Embodied Manipulation: A Bimanual-Mobile Robot Manipulation Dataset for Household Tasks}

\author[1]{\textbf{Tianle Zhang}}
\author[1*]{\textbf{Dongjiang Li}}
\author[1]{\textbf{Yihang Li}}
\author[1]{\textbf{Zecui Zeng}}
\author[1,2]{\textbf{Lin Zhao}}
\author[1]{\textbf{Lei Sun}}
\author[1]{\textbf{Yue Chen}}
\author[1]{\textbf{Xuelong Wei}}
\author[1]{\textbf{Yibing Zhan}}
\author[1]{\textbf{Lusong Li}}
\author[1]{\textbf{Xiaodong He}}
\affil[1]{JD Explore Academy, China}
\affil[2]{Beijing Institute of Technology, China}
\affil[ ]{\texttt{tianle-zhang@outlook.com, \{lidongjiang5, liyihang18, zengzecui1\}@jd.com, zhaolins@foxmail.com, \{sunlei155, chenyue21, weixuelong1\}@jd.com, zybjy@mail.ustc.edu.cn, lilusong@gmail.com, xiaodong.he@jd.com}}

\begin{document}

\maketitle

\let\oldthefootnote\thefootnote
\renewcommand{\thefootnote}{\fnsymbol{footnote}}
\footnotetext[1]{\phantomsection\hypertarget{myth}{Corresponding Author}}
\let\thefootnote\oldthefootnote

\begin{abstract}

The advancements in embodied AI are increasingly enabling robots to tackle complex real-world tasks, such as household manipulation. However, the deployment of robots in these environments remains constrained by the lack of comprehensive bimanual-mobile robot manipulation data that can be learned.
Existing datasets predominantly focus on single-arm manipulation tasks, while the few dual-arm datasets available often lack mobility features, task diversity, comprehensive sensor data, and robust evaluation metrics; they fail to capture the intricate and dynamic nature of household manipulation tasks that bimanual-mobile robots are expected to perform. To overcome these limitations, we propose \textbf{BRMData}, a \textbf{B}imanual-mobile \textbf{R}obot \textbf{M}anipulation \textbf{Data}set specifically designed for household applications. BRMData encompasses 10 diverse household tasks, including single-arm and dual-arm tasks, as well as both tabletop and mobile manipulations, utilizing multi-view and depth-sensing data information. Moreover, BRMData features tasks of increasing difficulty, ranging from single-object to multi-object grasping, non-interactive to human-robot interactive scenarios, and rigid-object to flexible-object manipulation, closely simulating real-world household applications. Additionally, we introduce a novel Manipulation Efficiency Score (MES) metric to evaluate both the precision and efficiency of robot manipulation methods in household tasks. We thoroughly evaluate and analyze the performance of advanced robot manipulation learning methods using our BRMData, aiming to drive the development of bimanual-mobile robot manipulation technologies. The dataset is now open-sourced and available at \url{https://embodiedrobot.github.io/}.

\end{abstract}

\section{Introduction}  
Recent advancements in large language models (LLMs) \cite{achiam2023gpt,touvron2023llama,taori2023stanford,chiang2023vicuna} and multimodal large language models (MLLMs) \cite{driess2023palm,zhu2023minigpt,liu2024visual} have significantly advanced the field of embodied manipulation, enhancing the capacity for robots to interact with the 3D physical world in a more human-like manner. This technology aims to endow robots with the ability to perform manipulation tasks, mirroring human dexterity and adaptability. In robot manipulation tasks, household environments recently attract significant attention from researchers due to the prevalence of human-operated tasks and the broad range of practical applications for robots \cite{xiong2023robotube,xiao2024robi}. Moreover, household tasks present a substantial challenge due to their diversity and the complex interactions they entail with various objects. To enable robots to autonomously and efficiently complete household tasks, it is crucial to train manipulation policies for bimanual-mobile robots that combine the manipulation of humanoid robots with autonomous mobility. A central ingredient for training such manipulation policies is diverse training datasets of bimanual-mobile robots. 

Recently, related studies have shown that comprehensive robot manipulation datasets are essential for developing and refining policy methods, enabling robots to perform complex household tasks with human-like dexterity \cite{padalkar2023open,iyer2024open,khazatsky2024droid,fu2024mobile}. However, existing datasets primarily focus on single-arm manipulation tasks, such as RT-1 \cite{brohan2022rt}, RH20T \cite{fang2023rh20t}, and RT-X \cite{padalkar2023open}, which are limited to basic object grasping and placement. This narrow focus fails to address the complexities and coordination demands of bimanual-mobile manipulations for household tasks. Compared with single-arm manipulation datasets, dual-arm manipulation datasets are rare. Although a few datasets include dual-arm manipulations, they often lack mobility features \cite{zhao2023learning} or are limited to simple sensory setups \cite{fu2024mobile}, lacking comprehensive sensor data, task variety,  progressive task difficulty, and robust evaluation metrics. These restrictions hamper their ability to comprehensively represent the complex tasks that bimanual-mobile robots must perform in diverse household environments. 
\begin{figure}[t]
    \centering
    \vspace{-0.3cm}
    \includegraphics[width=0.95\linewidth]{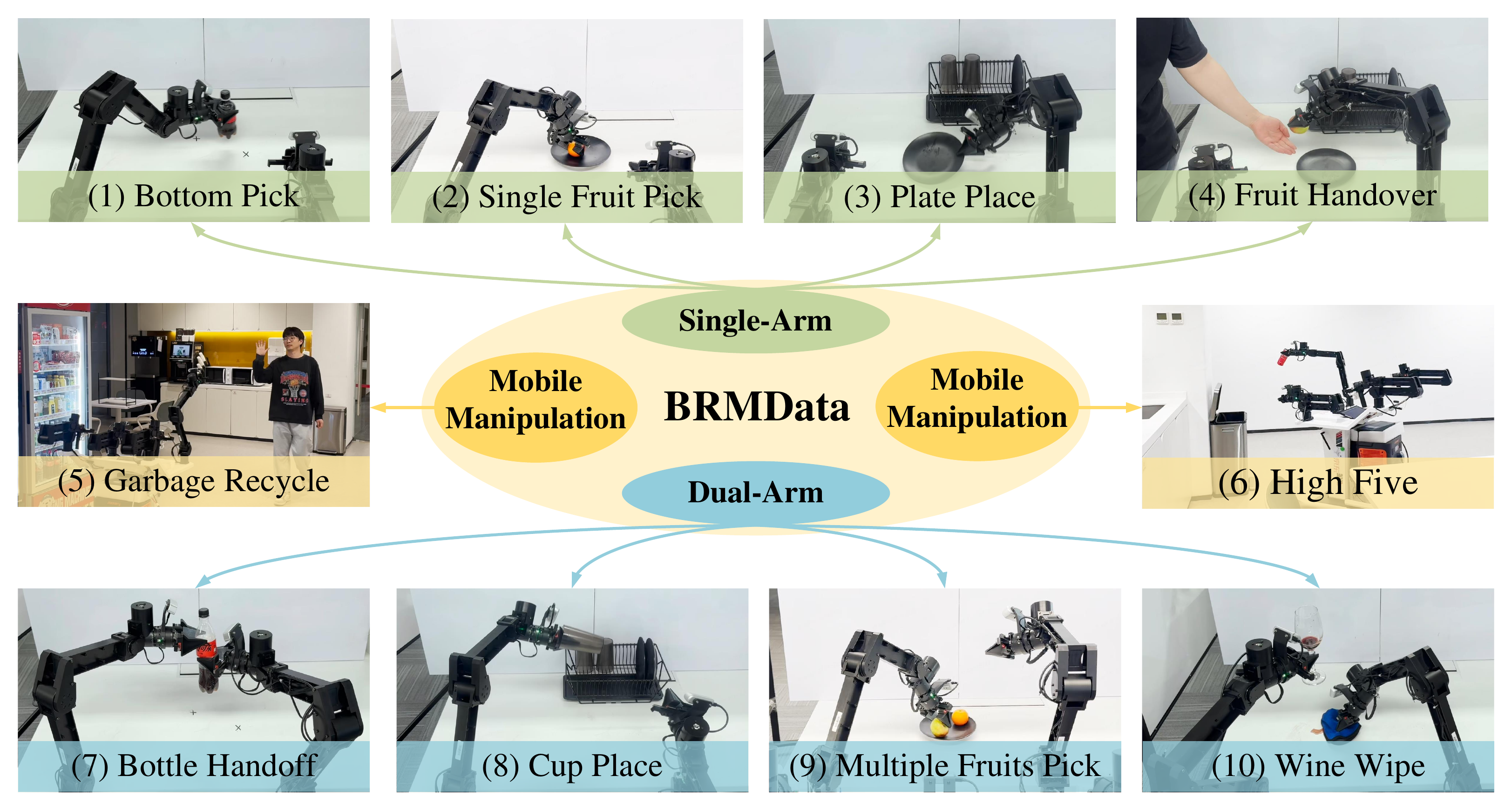}
    \caption{Illustration of our bimanual-mobile robot manipulation tasks.}
    \label{fig:tasks}
    \vspace{-1.0cm}
\end{figure}

Based on the aforementioned discussions, we propose a bimanual-mobile Robot Manipulation Dataset (BRMData). BRMData is a novel dataset meticulously designed to advance the development and evaluation of bimanual-mobile robots for household tasks. BRMData captures the intricacies of real-world tasks through human demonstrations, offering a true reflection of the dexterity and interaction complexity required. As depicted in Figure \ref{fig:tasks}, this dataset comprises 10 diverse manipulation tasks, including single-object pick-and-place manipulation, dual-arm coordination, human-robot interactions, and mobile manipulation. These tasks are specifically selected to challenge and develop robot manipulation capabilities incrementally. Moreover, the key properties of BRMData are emphasized as follows:

\textbf{Task Diversity:} BRMData encompasses a broad spectrum of manipulation tasks, including single-arm and dual-arm tasks, as well as tabletop and mobile manipulations. This diversity enhances the adaptability of robot manipulation learning methods across various real-world scenarios.

\textbf{Progressive Difficulty:} BRMData features tasks with escalating complexity, from single-object to multi-object grasping, non-interactive to human-robot interactive scenarios, and rigid-object to flexible-object manipulation. This progression facilitates the development of diverse robot manipulation learning methods.

\textbf{Sensory Data Comprehensiveness:} Employing RGBD camera sensors to perceive multi-view, static and dynamic imaging, depth-sensing and cross-sensor data, BRMData provides comprehensive and essential sensory information for testing the robustness of robot manipulation learning methods.

\textbf{Manipulation Efficiency Score:} We design a novel evaluation metric: Manipulation Efficiency Score (MES), which considers both the success rate and the efficiency of robot manipulation learning methods. This metric provides a more comprehensive evaluation, ensuring that developed manipulation algorithms not only complete tasks effectively but also promptly. 

Besides, based on MES, we further thoroughly evaluate the performance of the state-of-the-art robot manipulation learning methods through our BRMData, for promoting the iterative development of the methods.

\section{Related Work}

\subsection{Robot Manipulation Learning Methods}
In the realm of robot learning, the utilization of human demonstrations \cite{sharma2018multiple,bahl2023affordances,jang2022bc,lynch2023interactive,brohan2022rt} as a means to imbue robots with manipulation capabilities has garnered significant attention.
Robomimic \cite{mandlekar2021matters} conducts a comprehensive examination of six offline learning algorithms across various simulated and real-world manipulation tasks, delves into the nuances of learning from offline human demonstrations for robot manipulation. DayDreamer \cite{wu2023daydreamer}  enables robots to learn from minimal interaction by planning within a learned world model, which significantly reduces the trial-and-error process typically required in physical settings. 
CALVIN \cite{mees2022calvin} offers environments for policy learning under language instructions for long-horizon robotic manipulation tasks, emphasizing the importance of comprehensive datasets in enhancing robot-environment interactions. Similarly, DROID \cite{khazatsky2024droid} provides extensive data for robot manipulation, underscoring the necessity of testing and optimizing algorithms for dynamic, real-world applications.
Moreover, ARMBench \cite{mitash2023armbench} and PLEX \cite{thomas2023plex} have expanded the application scope by incorporating not only basic manipulation tasks but also more complex scenarios and diverse operational strategies. In addition, action chunking with transformers (ACT) \cite{zhao2023learning} learns a generative model over action sequences based on a conditional variational autoencoder \cite{sohn2015learning}.  Diffusion policy \cite{chi2023diffusion} is a novel approach to generating robot behavior by representing a robot’s visuomotor policy as a conditional denoising diffusion process.


\subsection{Robot Manipulation Datasets}
Robot manipulation datasets \cite{ebert2021bridge, fang2023rh20t, walke2023bridgedata, padalkar2023open} have seen significant evolution to support imitation learning and policy generalization in complex tabletop tasks and varied physical environments.
RH20T \cite{fang2023rh20t} and RoboAgent \cite{bharadhwaj2023roboagent} stand out as examples, enabling robots to learn and execute tasks effectively in both simulated and real settings.
The MIME dataset \cite{sharma2018multiple} utilizes multiple collection techniques, including kinesthetic teaching and visual demonstrations, to gather data on human-robot interactions. 
Trained on a substantial dataset of over 130k episodes spanning 17 months, RT-1 \cite{brohan2022rt} leverages transformer-based architectures to facilitate real-world control tasks at scale.
While BridgeData V2 \cite{walke2023bridgedata} and RH20T \cite{fang2023rh20t} offer depth-sensing and multi-view capabilities, they are limited in movable capabilities and restricted to single-arm manipulation tasks. 
In contrast, our proposed dataset, BRMData, integrates cross-sensor and depth-sensing features, distinguishing it from similar datasets like ALOHA \cite{zhao2023learning} and Mobile ALOHA \cite{fu2024mobile} lacking such capabilities.


\section{BRMData}

In this section, we comprehensively introduce the proposed dataset, called Bimanual-mobile Robot Manipulation Dataset (BRMData), to the community. Firstly, the data collection setup is presented. Then, the properties of BRMData are emphasized. Finally, dataset discussions are given.

\subsection{Data Collection Setup}

\textbf{1) Robot Platform:} As illustrated in Figure \ref{fig:hardware_architecture}, the data collection hardware platform for BRMData comprises 4 ARX-5 robot arms, each equipped with 7 joints and a parallel gripper. The \textit{Follower-1} and \textit{Follower-2} robot arms, positioned at the front of the platform, are specifically engineered for advanced bimanual manipulations. Each of both arms features a movable RGBD camera sensor at the wrist. The flexibility afforded by these cameras facilitates variable viewing angles and provides detailed visual feedback, crucial for precision in manipulation tasks.  Positioned centrally between \textit{Follower-1} and \textit{Follower-2} is a fixed RGBD camera sensor. This middle camera provides a consistent and broad visual reference across the manipulation area. 

\begin{wrapfigure}{r}{0.5\textwidth}
    \centering
    \includegraphics[width=1.0\linewidth]{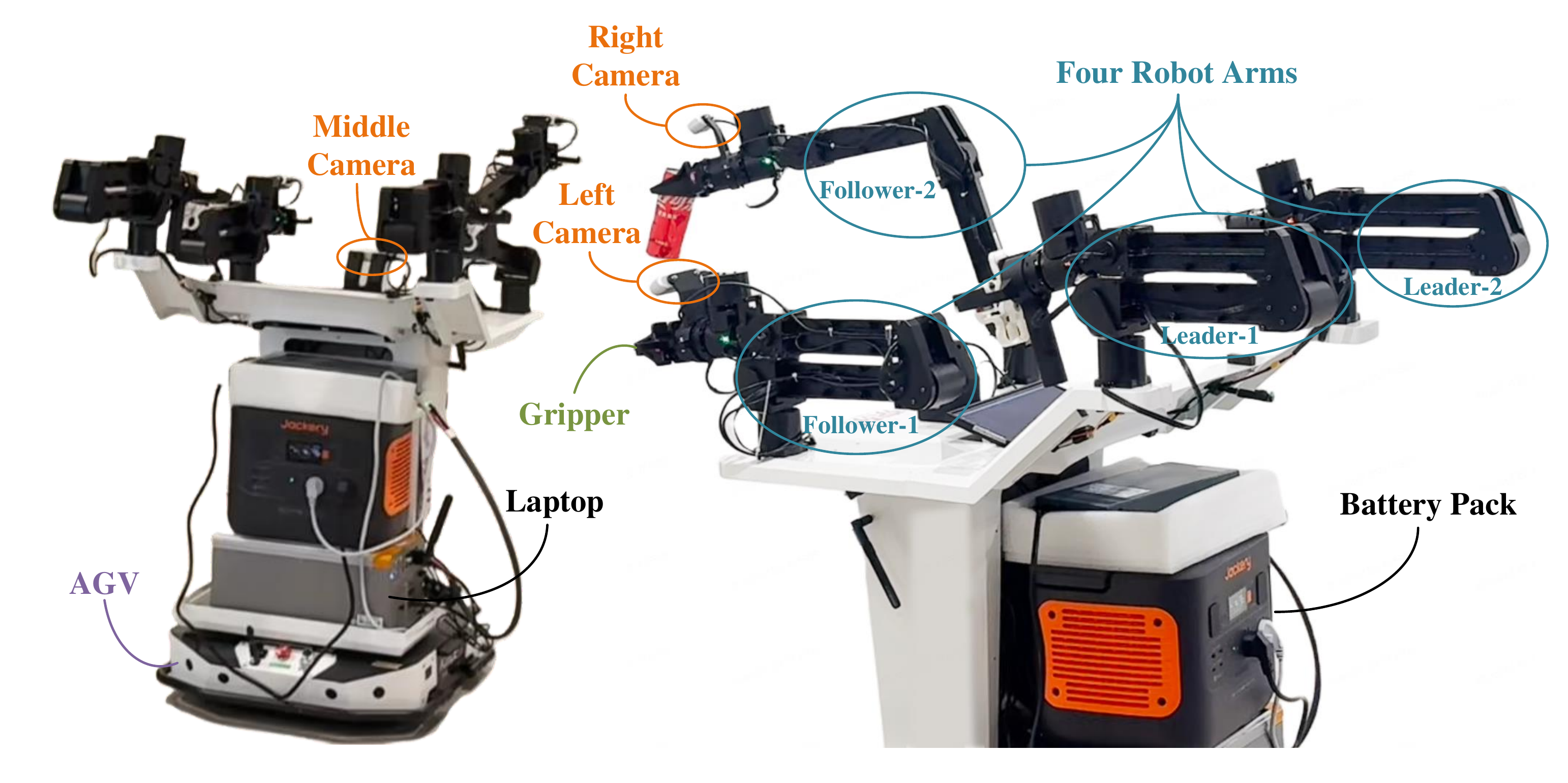}
    \caption{Illustration of robot platform.}
    \label{fig:hardware_architecture}
\end{wrapfigure}


The \textit{Leader-1} and \textit{Leader-2} robot arms, located at the rear of the platform, are designated for teleoperation by human experts. This configuration facilitates the synchronous replication of human-directed actions from the leader arms to the follower arms. The platform includes an automated guided vehicle equipped with a two-wheel differential Tracer mobile base for providing essential mobility.
The computational demands of the platform are managed by an industrial computer equipped with an NVIDIA 4090 graphics card, an Intel i7-13700 CPU, and 32GB of RAM. This computing setup ensures efficient camera data processing and supports real-time algorithms necessary for effective manipulation and interaction.



\begin{figure}[h]
  \centering
  \begin{minipage}{.5\textwidth}
    \centering
    \includegraphics[width=\linewidth]{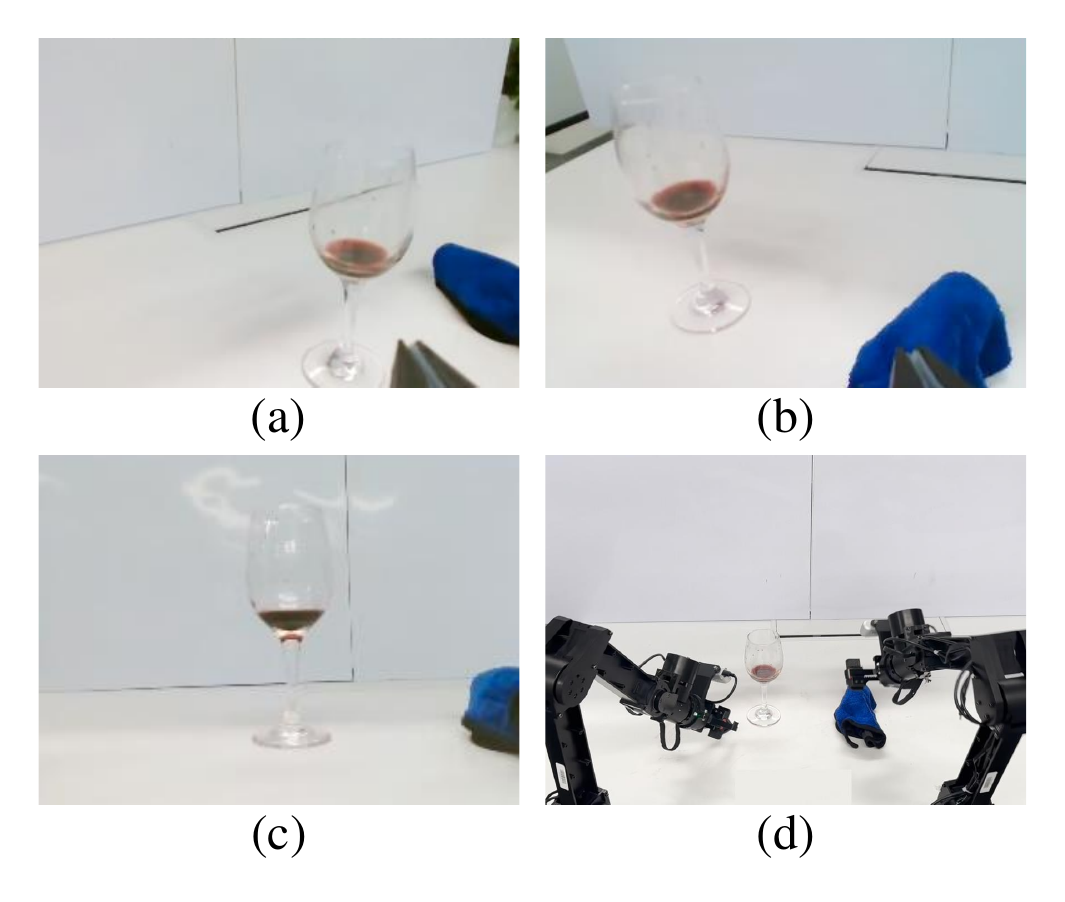}
  \end{minipage}%
  \begin{minipage}{.5\textwidth}
    \centering
    \caption{Illustration of the multi-view data collection of BRMData. (a) \textbf{Left Image}: Captured from the left wrist-mounted camera, offering detailed visual input from the left-arm's side. (b) \textbf{Right Image}: Obtained from the right wrist-mounted camera, providing critical visual information from the robot's right side. (c) \textbf{Middle Image}: Sourced from the centrally mounted fixed camera, presenting a broad and central perspective of the operational area. (d) \textbf{Human Perspective View}: Captured from a human viewpoint, not through the robot platform, giving an external perspective on the scene.}
    \label{fig:multi_view}
  \end{minipage}
\end{figure}

\textbf{2) Data Collection Protocol:} The data collection protocol for BRMData is meticulously designed to capture a wide array of manipulation tasks comprehensively. 
Specifically, as shown in Figure \ref{fig:multi_view}, both the left and right wrist-mounted RGBD cameras and the middle camera record RGB and depth images at a resolution of $640 \times 480$. The integration of these three views provides a multi-perspective visual input, essential for constructing a detailed 3D representation of the operational environment. Meanwhile, to bring in cross-sensor data, for \textit{Bottle Pick} and \textit{Bottle Handoff} tasks, image data are captured using Orbbec Dabai cameras at 30Hz. For other manipulation tasks, Intel Realsense D435 cameras are used to capture image data at 60Hz. In addition, each manipulation task within the dataset includes the collection of 50 action trajectories, gathered at intervals that are specifically tailored to the complexity and specific requirements of each task scenario. To enrich the behavior diversity of the dataset, each manipulation task is performed by three human experts. These experts use the \textit{Leader-1} and \textit{Leader-2} robot arms to demonstrate the manipulation tasks, with their actions being synchronously replicated on the \textit{Follower-1} and \textit{Follower-2} arms. 
This structured approach ensures that every trajectory is unique, capturing a broad spectrum of challenges and demands encountered in real-world bimanual-mobile robot manipulation.

\subsection{Properties of BRMData}
BRMData is structured around diverse manipulation tasks. For detailed task definitions and step-by-step descriptions, Please refer to Appendix \ref{appendix:tasks}. Besides, we further emphasize the following properties of our BRMData.

\subsubsection{Task Diversity}
BRMData incorporates a broad spectrum of robotic manipulation tasks that extend from single-arm to dual-arm manipulations, and tabletop to mobile manipulations, as detailed in Table \ref{task_types}. The dataset provides a foundation for investigating fundamental robotic interactions like \textit{Bottle Pick}, \textit{Plate Place}, and \textit{Single Fruit Pick}. These tasks encompass basic object handling as well as advanced human-robot interactions. Tasks such as \textit{Bottle Handoff} and \textit{Wine Wipe} further the complexity by necessitating coordinated bimanual actions and managing multi-object interactions, representing advanced manipulation challenges. The mobile manipulation tasks, including \textit{Garbage Recycle} and \textit{High Five}, introduce navigation and dynamic interaction capabilities, essential for operational efficacy in changing environments. The diverse task range in BRMData enables comprehensive testing and refinement of robotic manipulation strategies suitable for a variety of applications.

\begin{table}[h]
\renewcommand\arraystretch{1.2}
\centering
\caption{Key attribute analysis of robot manipulation tasks across BRMData}
\resizebox{0.95\textwidth}{!}{
\begin{tabular}{ccccccc}
\toprule
Type & Task              & Movable       & \makecell{Human-robot \\  Interaction} & \makecell{Multi-object \\ Interaction} & \makecell{Object \\ Attribute} & Steps \\ \hline
\multirow{4}{*}{\makecell{Tabletop Single-arm \\ Manipulation}} & Bottle Pick     & $\times$     & $\times$ & $\times$ & Rigid & 400  \\
     & Single Fruit Pick & $\times$     & $\times$          & $\checkmark$ & Rigid & 400   \\
     & Plate Place       & $\times$     & $\times$          & $\times$     & Rigid & 600   \\
     & Fruit Handover    & $\times$     & $\checkmark$      & $\times$     & Rigid & 500   \\ \hline
\multirow{4}{*}{\makecell{Tabletop Dual-arm \\ Manipulation}}   & Bottle Handoff  & $\times$     & $\times$ & $\times$ & Rigid & 800  \\
     & Cup Place         & $\times$     & $\times$          & $\times$     & Rigid & 800   \\
     & Multi Fruit Pick  & $\times$     & $\times$          & $\checkmark$ & Rigid & 900   \\
     & Wine Wipe         & $\times$     & $\times$          & $\checkmark$ & Flexible  & 1000  \\ \hline
\multirow{2}{*}{Mobile Manipulation}              & Garbage Recycle & $\checkmark$ & $\times$ & $\times$ & Rigid & 1200 \\
     & High Five         & $\checkmark$ & $\checkmark$      & $\times$     & Flexible & 500   \\ \bottomrule
\end{tabular}}
\label{task_types}
\end{table}

\subsubsection{Progressive Difficulty}
BRMData is carefully designed to present a series of tasks with increasing levels of complexity and diversity, not only in the manipulation tasks but also in the attributes and interaction with objects. This design facilitates a gradual elevation in the skillset required from the robots, focusing on their adaptability and precision in handling different object attributes and interaction complexities.

\textbf{From Single-Object to Multi-Object: } Initially, tasks such as \textit{Bottle Pick} and \textit{Single Fruit Pick} involve interactions with a single object, allowing for focused assessments of basic manipulation capabilities. However, as the complexity level rises, the tasks evolve to include multi-object interactions. Tasks like \textit{Multi Fruit Pick} and \textit{Wine Wipe} require the robot to manage multiple objects either sequentially or simultaneously. This shift from single to multi-object manipulation tests the robot’s capability in task planning, object prioritization, and execution of complex sequences involving various objects.

\textbf{From Non-Interaction to Human-Robot Interaction:} BRMData includes tasks without or with human-robot interactions, as shown in Table \ref{task_types}. Compared with tasks without human-robot interactions, tasks with human-robot interactions (e.g., \textit{Fruit Handover} and \textit{High Five}) require a robot to cooperate with humans to achieve the tasks. From the robot perception perspective, human actions are usually different every time, despite the same behavior. This adds a certain amount of random disturbance to the task itself, increasing the difficulty of the task. 

\textbf{From Rigid-Object to Flexible-Object: } BRMData differentiates between rigid and flexible objects. Initial tasks like \textit{Bottle Pick}, \textit{Single Fruit Pick}, and \textit{Plate Place} involve rigid objects, which are generally simpler to manipulate due to their stable and predictable nature. These tasks serve as fundamental tests of a robot's capabilities in gripping, object recognition, and precise placement, laying the groundwork for more complex manipulations. 
As the tasks progress in complexity, the dataset introduces flexible objects, as seen in \textit{Wine Wipe} and \textit{High Five} tasks. 
The manipulation of flexible objects poses additional challenges due to their deformable nature, which requires more sophisticated sensory and feedback mechanisms to adjust grip strength and manipulation strategies dynamically. For instance, in the \textit{Wine Wipe} task, the robot must not only lift a wine glass but also clean a spill with a cloth, involving precise movements and pressure control to avoid slippage or damage.
The \textit{High Five} task exemplifies the complexity of interacting with human hands, which introduces unpredictability in object behavior and demands a nuanced understanding of dynamic material properties. 
This progression from rigid to flexible object manipulation significantly raises the difficulty level, challenging the robot's capacity to handle a diverse range of material properties and interaction scenarios.

\subsubsection{Sensory Data Comprehensiveness}
\textbf{Static and Dynamic Imaging: } BRMData utilizes a combination of two mobile RGBD cameras mounted on the robot arms and a centrally positioned fixed camera to ensure an exhaustive visual representation of the manipulation environment. The mobile cameras provide changing perspectives that track the motion of the robot arms, capturing detailed data critical for analyzing interactions in real-time. The fixed camera maintains a consistent view of the entire scene, enabling the capture of invariant spatial references. This multi-view arrangement significantly enhances the dataset by supplying comprehensive spatial and temporal insights, essential for the complex tasks being performed.

\textbf{Depth-Sensing: }We utilize RGBD cameras to acquire spatial information of the physical environment. These sensors facilitate the 3D environmental perception by providing depth information that complements the RGB imagery. This dual modality is essential for precise object detection, spatial reasoning, and execution of complex manipulative tasks, particularly in environments where object depth and positioning are critical for successful interaction.

\textbf{Cross-Sensor: } Our BRMData incorporates data collected by different types of cameras. For specific tasks such as the \textit{Bottle Pick} and \textit{Bottle Handoff}, image data are captured using Orbbec Dabai cameras operating at a frequency of 30Hz. For more dynamic manipulation tasks, Intel Realsense D435 cameras, which capture data at a higher frequency of 60Hz, are employed to accommodate the quick movements and frequent data updates required. These cross-sensor data enrich our dataset and provide a possibility to test the adaptability of robot manipulation learning methods to different sensing devices.

\subsubsection{Manipulation Efficiency Score}
Evaluation metrics are crucial to measuring the performance of the robot manipulation learning methods. Success rate (SR) is generally used to evaluate the performance of the methods in robot manipulation tasks. While SR provides a measure of task completion success, it does not account for the efficiency of task execution. High SR may still be accompanied by long execution times, which is impractical for real-world applications where time efficiency is crucial. Therefore, we introduce a novel metric: \textbf{Manipulation Efficiency Score (MES)}. It is defined as $\text{MES} = 100\frac{\text{SR}}{T}$, where $T$ denotes the average execution time. This metric captures both the effectiveness and efficiency of robot manipulation methods, rewarding those that achieve high success rates with minimal execution time, thereby highlighting both the quality and speed of task execution.




\subsection{Dataset Discussions}
To further analyze the highlights of our BRMData, we compare and discuss our dataset with existing datasets, as shown in Table \ref{compare1}. According to the comparison, BRMData stands out due to its comprehensive coverage of bimanual-mobile manipulations. Through the integration of dual-arm coordination, diverse task settings, and comprehensive sensor data collection, BRMData offers an enriched training environment unparalleled by existing representative datasets. Specifically, compared with the other datasets, BRMData incorporates not only human-robot tasks but also cross-sensor tasks, along with depth-sensing information. This enhancement mitigates previous limitations and empowers robots to operate with heightened versatility and adaptability, particularly in dynamically changing settings. In addition, BRMData has more unique properties than the other datasets: task diversity, progressive difficulty, and sensory data comprehensiveness, which can greatly facilitate research in robot manipulation learning.


\begin{table}[h]
\centering
\caption{Comparison to existing datasets for robot manipulation.}
\resizebox{1.0\textwidth}{!}{
\begin{tabular}{@{}ccccccccc@{}}
\toprule
Dataset                  & \makecell{Human \\ Demonstration} & Movable &  Dual-arm & Depth-sensing & Multi-view & Torque & \makecell{Human-Robot \\ Interaction} & Cross-Sensor\\ \midrule
MIME \cite{sharma2018multiple}     & $\checkmark$ & $\times$    & $\checkmark$    & $\checkmark$  & $\checkmark$  &  $\times$ & $\times$ & $\times$                \\
RT-1  \cite{brohan2022rt}         & $\checkmark$ & $\checkmark$    & $\times$    & $\times$    &  $\times$  &  $\times$ & $\times$ & $\times$                \\
RH20T  \cite{fang2023rh20t}          &  $\checkmark$    & $\times$   & $\times$   & $\checkmark$     &  $\checkmark$ &  $\times$ & $\times$ & $\times$                \\
BridgeData V2  \cite{walke2023bridgedata}        &   $\times$   & $\times$   & $\times$   & $\checkmark$   & $\checkmark$  &  $\times$ & $\times$ & $\times$                  \\
ALOHA   \cite{zhao2023learning}          &   $\checkmark$   & $\times$   & $\checkmark$   & $\times$ &  $\checkmark$ &  $\checkmark$ & $\times$ & $\times$                    \\
Mobile ALOHA  \cite{fu2024mobile}          &   $\checkmark$   & $\checkmark$   & $\checkmark$   & $\times$  &  $\checkmark$ &  $\checkmark$ & $\checkmark$ & $\times$                    \\
\textbf{BRMData (Ours)}      & $\checkmark$ & $\checkmark$    & $\checkmark$    &  $\checkmark$  &  $\checkmark$   &  $\checkmark$  & $\checkmark$ &$\checkmark$                   \\ \bottomrule
\end{tabular}
}
\label{compare1}
\end{table}





\section{Experiments}
Rather than delving into an exhaustive analysis of different policy network design choices, the goal of our experiments is to explore the extent to which our dataset evaluates and enhances the application of robot manipulation learning methods in executing household tasks. Our experiments are designed to address the following key questions: (a) \textit{Is BRMData effective for testing various robot manipulation learning methods?} (b) \textit{How do the properties of BRMData support the evaluation of methods?}

\subsection{Experimental Setup}
To investigate the performance and adaptability of various robot manipulation learning methods on BRMData, we conduct repetitive experiments using both single-task and multi-task methods. The methods used in the single-task setting include Action Chunking with Transformers (ACT) \cite{zhao2023learning} and Diffusion Policy (DP) \cite{chi2023diffusion}. The methods used in the multi-task experiments include MT-ACT \cite{nair2023r3m}, MT-ACT-EB3, and MT-ACT-R3M. Additional details of the experimental setup and the methods can be found in Appendix \ref{appendix:exp_setup} and \ref{appendix:robot_learning_methods}.





\subsection{Experimental Results}


\subsubsection{Is BRMData effective for testing various robot manipulation learning methods?}
\begin{table}[h]
\centering
\renewcommand{\arraystretch}{1.1} 
\caption{Testing results of different robot manipulation learning methods.}
\resizebox{0.95\textwidth}{!}{
\begin{tabular}{cccccccc}
\toprule
\multicolumn{2}{c}{\multirow{2}{*}{Method}} & \multicolumn{2}{c}{Multiple Fruits Pick} & \multicolumn{2}{c}{Cup Place} & \multicolumn{2}{c}{Garbage Recycle} \\
\multicolumn{2}{c}{}                      & SR          & MES           & SR           & MES           & SR          & MES           \\ \midrule
\multirow{2}{*}{Single-task} & ACT        & 60          & 3.61          & 90           & \textbf{6.06} & 55          & \textbf{2.62} \\
                             & DP         & 80          & 2.94          & 80           & 4.20          & 40          & 1.57          \\ \cline{2-8} 
\multirow{3}{*}{Multi-task}  & MT-ACT     & \textbf{90} & \textbf{4.78} & 90           & 5.56          & \textbf{70} & 2.50          \\
                             & MT-ACT-EB3 & 70          & 2.08          & \textbf{100} & 3.57          & 40          & 1.03          \\
                             & MT-ACT-R3M & 50          & 2.79          & 90           & 5.82          & -           & -             \\ \bottomrule
\end{tabular}
}
\label{sin_mul}
\end{table}
To effectively test various robot manipulation learning methods on BRMData, we select three representative robot manipulation tasks: \textit{Multiple Fruits Pick} (tabletop dual-arm manipulation), \textit{Cup Place} (tabletop dual-arm interaction manipulation), \textit{Garbage Recycle} (mobile manipulation). The testing results are shown in Table \ref{sin_mul}.
In single-task method experiments, the ACT method generally outperforms the DP method in terms of execution efficiency. Despite a lower SR in the \textit{Multiple Fruits Pick} task, ACT achieves a superior MES of 3.61, indicating better efficiency. In the \textit{Cup Place} task, ACT excels with a 90\% SR and the highest MES of 6.06, highlighting its precision and speed in task execution. This is because the denoising process of DP is time-consuming. Meanwhile, the experiments find that DP is good at diffusion learning of action sequences, but is not particularly sensitive to visual states, compared with ACT. Moreover, the generalization of both ACT and DP is weak, particularly when there is a perturbation in the object's position, which often results in an inability to complete the task. 

In multi-task method experiments, MT-ACT proves to be highly adaptable and efficient, achieving a 90\% SR and an MES of 4.78 in the \textit{Multiple Fruits Pick} task, along with strong performance across other tasks. On the contrary, MT-ACT-R3M uses a pre-trained R3M model but still performs poorly because it is limited by the smaller image size input, resulting in the loss of visual information. Compared to MT-ACT, MT-ACT-EB3 loses visual information, has larger model parameters, and requires longer inference time. The experimental results show that different robot manipulation methods can be tested on our dataset. The possible users can test both single-task learning methods and multi-task learning methods and make a comparison with the existing methods. Besides, to further assess the robustness of the learning methods under varying conditions, additional experiments are conducted as detailed in Appendix \ref{appendix:robustness_test}.

\subsubsection{How do the properties of BRMData support the evaluation of methods?}
The tasks of BRMData have the properties of task diversity, progressive difficulty, and sensory data comprehensiveness. They can facilitate the verification of method effectiveness across diverse tasks. Both the basic capability to finish simple tasks and the upper boundary for complex tasks of the methods can be illustrated. 

\begin{figure}[h]
    \centering
    \includegraphics[width=1.0\linewidth]{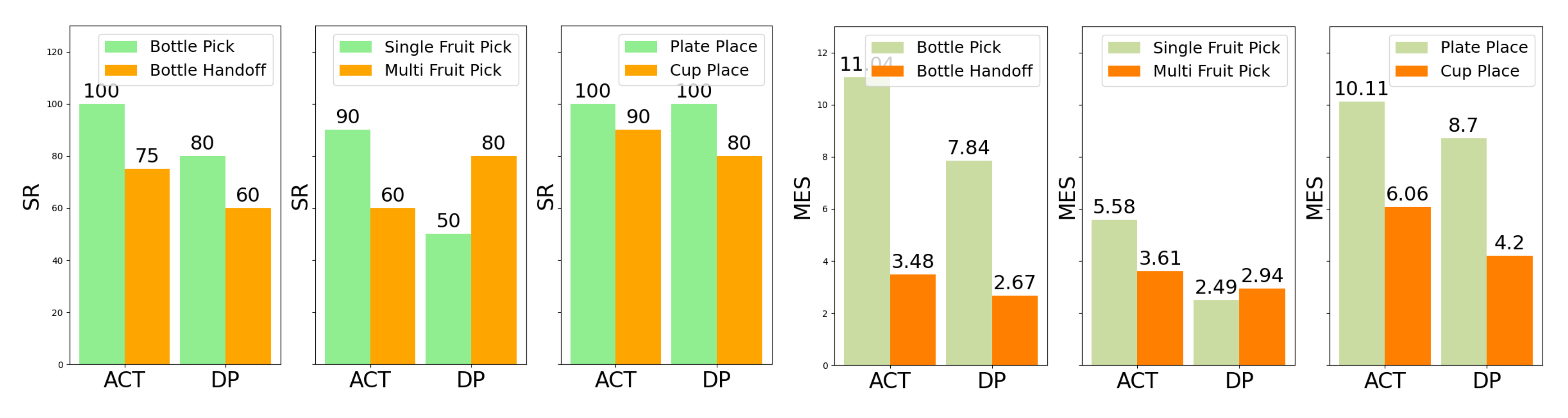}
    \caption{Performance comparison in single-arm and dual-arm robot manipulation tasks.}
    \label{table_diff_1}
\end{figure}

\textbf{Verification of method effectiveness on diverse tasks.}
As illustrated in Figure \ref{table_diff_1},  the experimental outcomes from single-task methodologies, namely ACT and DP, exhibit a spectrum of success and operational efficiency, thereby offering a robust validation of the evaluative methods employed. Within the ambit of single-arm operations, such as \textit{Bottle Pick}, \textit{Single Fruit Pick}, and \textit{Bottle Place}, the ACT approach registers an SR between 90\% and 100\%, and an MES ranging from 5.58 to 11.04. Conversely, the DP strategy secures an SR of 50\% to 90\%, with MES values spanning 2.49 to 8.7. In dual-arm tasks, the ACT method attains a SR of 60\% to 90\%, alongside an MES between 3.48 and 6.06, whereas DP records a SR of 60\% to 80\%, and an MES ranging from 2.94 to 4.2. The testing results demonstrate that the ACT method leads to a relatively higher success rate and higher efficiency. The application across a varied set of tasks on BRMData substantiates the effectiveness of the manipulation methods under investigation.

\begin{wraptable}{r}{0.53\textwidth}
\centering
\caption{Testing results in robot manipulation tasks without and with human-robot interaction.}
\begin{tabular}{ccccccc}
\toprule
\multirow{2}{*}{Method} & \multicolumn{2}{c}{Single Fruit Pick} & & \multicolumn{2}{c}{Fruit Handover} \\\arrayrulecolor{gray}\cline{2-3} \cline{5-6} \arrayrulecolor{black} 
    & SR  & MES &   & SR & MES  \\ \midrule
ACT & \textbf{90}  & \textbf{5.58} & & \textbf{85} & \textbf{8.92} \\
DP  & 50  & 2.49  & & 40 & 3.56  \\ \bottomrule
\end{tabular}
\label{table_diff_2}
\end{wraptable}


\textbf{Illustration of method capability boundaries on tasks with progressive difficulty.}
As illustrated in Table \ref{table_diff_2}, the tasks of \textit{Single Fruit Pick} and \textit{Fruit Handover} involve the picking and placing of a single fruit, differing only in the occurrence of human-robot interaction. ACT method achieves an SR of 90\% and an MES of 5.58, whereas the DP method records an SR of 50\% and a MES of 2.49. These results indicate that the ACT method performs effectively in basic tasks, while the DP method exhibits moderate performance. In the more challenging \textit{Fruit Handover} task, the performance of both methods declines, with ACT achieving an 85\% SR and an 8.92 MES, and DP a 40\% SR and a 3.56 MES. To be noticed, the task \textit{Single Fruit Pick} involves repeating the picking and placing twice time, thus resulting in a small MES. This comparative analysis suggests that the ACT method not only maintains superior task execution in complex scenarios but also hints at a potentially higher threshold of capability. Conversely, the DP method's diminished performance in the \textit{Fruit Handover} task suggests it may have reached its capability limit for tasks of increased difficulty. In addition, the impact of mobility on task complexity is discussed in Appendix \ref{Impact of Mobility on Task Complexity}.
\textbf{Effectiveness analysis of the MES metric for evaluating methods.} Table \ref{single_task_table} shows that the MES metric offers a comprehensive and fair measure of performance. By incorporating both SR and execution time, MES effectively penalizes methods that take longer to complete tasks, ensuring a balanced evaluation. For example, in the \textit{Multiple Fruits Pick} task, despite ACT having a lower SR (60\%) compared to DP (80\%), ACT’s MES of 3.61 is higher than DP’s 2.94. This highlights ACT's greater efficiency, demonstrating that MES can fairly assess the trade-offs between success rates and execution efficiency. Such a balance is crucial for real-world applications where both effectiveness and efficiency are essential.
The MES also excels in highlighting the trade-offs between different approaches. For instance, in the \textit{High Five} task, ACT achieves a high MES of 10.28, indicating both a high success rate and efficient execution. Conversely, DP achieves a lower MES of 4.86, pointing to longer execution times despite a reasonable SR of 80\%. This shows that MES can effectively differentiate between methods based on both their accuracy and operational efficiency. Furthermore, MES helps to identify inefficiencies in more complex tasks, such as \textit{Garbage Recycle}, where the need for both manipulation and mobility challenges the methods, resulting in lower MES values for both ACT (2.62) and DP (1.57).

\begin{table}[h]
\centering
\renewcommand{\arraystretch}{1.5} 
\caption{Testing results of different single-task models.}
\setlength\tabcolsep{1.5pt}
\resizebox{0.98\textwidth}{!}{
\begin{tabular}{@{}p{35pt}<{\centering}p{35pt}<{\centering}p{30pt}<{\centering}p{35pt}<{\centering}p{30pt}<{\centering}p{30pt}<{\centering}p{30pt}<{\centering}p{30pt}<{\centering}p{40pt}<{\centering}p{30pt}<{\centering}p{30pt}<{\centering}p{35pt}<{\centering}@{}}
\toprule
& Method & Bottle Pick & Bottle Handoff & Single Fruit Pick & Multiple Fruits Pick & Plate Place & Cup Place & Fruit Handover & High Five & Wine Wipe & Garbage Recycle \\ \midrule
\multirow{2}{*}{SR (\%)} & ACT & 100 & 75 & 90 & 60 & 100 & 90 & 85 & 100 & 100 & 55  \\
& DP &  80 & 60 & 50 & 80 & 100 & 80 & 40 & 80 & 80 & 40  \\ \cline{2-12}
\multirow{2}{*}{MES} & ACT & 11.04 & 3.48 & 5.58 & 3.61 & 10.11 & 6.06 & 8.92 & 10.28 & 5.06 & 2.62  \\
& DP & 7.84 & 2.67 & 2.49 & 2.94 & 8.70 & 4.20 &  3.56 & 4.86 & 3.38 & 1.57 \\ \bottomrule
\end{tabular}}
\smallskip
\label{single_task_table}
\end{table}

\section{Conclusion}
In this paper, BRMData is proposed to address the critical shortage of comprehensive datasets for dual-arm mobile robot manipulation, which represents a significant advancement in the field of embodied manipulation. By encompassing a range of bimanual and mobile manipulations, BRMData facilitates the exploration of complex interactions within dynamic environments that closely mimic real-world scenarios. This dataset not only supports the deployment of humanoid robots with enhanced capabilities but also provides a benchmark for testing and improving state-of-the-art models in robot manipulation. The diverse task configurations included in BRMData have proven instrumental in evaluating the adaptability and effectiveness of various robot manipulation learning methods. This has allowed for a detailed assessment of how different approaches perform under complex and varied conditions, thus contributing valuable insights into the development of more versatile and capable robotic systems. This work underscores the importance of such datasets in advancing our understanding of robotic capabilities and limitations, highlighting key areas for future exploration and development in the realm of embodied robots.

\clearpage

\bibliographystyle{unsrtnat}
\bibliography{references.bib}

\begin{thebibliography}{37}
\providecommand{\natexlab}[1]{#1}
\providecommand{\url}[1]{\texttt{#1}}
\expandafter\ifx\csname urlstyle\endcsname\relax
  \providecommand{\doi}[1]{doi: #1}\else
  \providecommand{\doi}{doi: \begingroup \urlstyle{rm}\Url}\fi

\bibitem[Achiam et~al.(2023)Achiam, Adler, Agarwal, Ahmad, Akkaya, Aleman, Almeida, Altenschmidt, Altman, Anadkat, et~al.]{achiam2023gpt}
Josh Achiam, Steven Adler, Sandhini Agarwal, Lama Ahmad, Ilge Akkaya, Florencia~Leoni Aleman, Diogo Almeida, Janko Altenschmidt, Sam Altman, Shyamal Anadkat, et~al.
\newblock Gpt-4 technical report.
\newblock \emph{arXiv preprint arXiv:2303.08774}, 2023.

\bibitem[Touvron et~al.(2023)Touvron, Lavril, Izacard, Martinet, Lachaux, Lacroix, Rozi{\`e}re, Goyal, Hambro, Azhar, et~al.]{touvron2023llama}
Hugo Touvron, Thibaut Lavril, Gautier Izacard, Xavier Martinet, Marie-Anne Lachaux, Timoth{\'e}e Lacroix, Baptiste Rozi{\`e}re, Naman Goyal, Eric Hambro, Faisal Azhar, et~al.
\newblock Llama: Open and efficient foundation language models.
\newblock \emph{arXiv preprint arXiv:2302.13971}, 2023.

\bibitem[Taori et~al.(2023)Taori, Gulrajani, Zhang, Dubois, Li, Guestrin, Liang, and Hashimoto]{taori2023stanford}
Rohan Taori, Ishaan Gulrajani, Tianyi Zhang, Yann Dubois, Xuechen Li, Carlos Guestrin, Percy Liang, and Tatsunori~B Hashimoto.
\newblock Stanford alpaca: An instruction-following llama model, 2023.

\bibitem[Chiang et~al.(2023)Chiang, Li, Lin, Sheng, Wu, Zhang, Zheng, Zhuang, Zhuang, Gonzalez, et~al.]{chiang2023vicuna}
Wei-Lin Chiang, Zhuohan Li, Zi~Lin, Ying Sheng, Zhanghao Wu, Hao Zhang, Lianmin Zheng, Siyuan Zhuang, Yonghao Zhuang, Joseph~E Gonzalez, et~al.
\newblock Vicuna: An open-source chatbot impressing gpt-4 with 90\%* chatgpt quality.
\newblock \emph{See https://vicuna. lmsys. org (accessed 14 April 2023)}, 2\penalty0 (3):\penalty0 6, 2023.

\bibitem[Driess et~al.(2023)Driess, Xia, Sajjadi, Lynch, Chowdhery, Ichter, Wahid, Tompson, Vuong, Yu, et~al.]{driess2023palm}
Danny Driess, Fei Xia, Mehdi~SM Sajjadi, Corey Lynch, Aakanksha Chowdhery, Brian Ichter, Ayzaan Wahid, Jonathan Tompson, Quan Vuong, Tianhe Yu, et~al.
\newblock Palm-e: An embodied multimodal language model.
\newblock In \emph{International Conference on Machine Learning}, pages 8469--8488. PMLR, 2023.

\bibitem[Zhu et~al.(2023)Zhu, Chen, Shen, Li, and Elhoseiny]{zhu2023minigpt}
Deyao Zhu, Jun Chen, Xiaoqian Shen, Xiang Li, and Mohamed Elhoseiny.
\newblock Minigpt-4: Enhancing vision-language understanding with advanced large language models.
\newblock In \emph{The Twelfth International Conference on Learning Representations}, 2023.

\bibitem[Liu et~al.(2024)Liu, Li, Wu, and Lee]{liu2024visual}
Haotian Liu, Chunyuan Li, Qingyang Wu, and Yong~Jae Lee.
\newblock Visual instruction tuning.
\newblock \emph{Advances in neural information processing systems}, 36, 2024.

\bibitem[Xiong et~al.(2023)Xiong, Fu, Zhang, Bao, Zhang, Huang, Xu, Garg, and Lu]{xiong2023robotube}
Haoyu Xiong, Haoyuan Fu, Jieyi Zhang, Chen Bao, Qiang Zhang, Yongxi Huang, Wenqiang Xu, Animesh Garg, and Cewu Lu.
\newblock Robotube: Learning household manipulation from human videos with simulated twin environments.
\newblock In \emph{Conference on Robot Learning}, pages 1--10. PMLR, 2023.

\bibitem[Xiao et~al.()Xiao, Gupta, Deng, Li, and Hsu]{xiao2024robi}
Anxing Xiao, Anshul Gupta, Yuhong Deng, Kaixin Li, and David Hsu.
\newblock Robi butler: Multimodal remote interaction with household robotic assistants.
\newblock In \emph{2nd Workshop on Mobile Manipulation and Embodied Intelligence at ICRA 2024}.

\bibitem[Padalkar et~al.(2023)Padalkar, Pooley, Jain, Bewley, Herzog, Irpan, Khazatsky, Rai, Singh, Brohan, et~al.]{padalkar2023open}
Abhishek Padalkar, Acorn Pooley, Ajinkya Jain, Alex Bewley, Alex Herzog, Alex Irpan, Alexander Khazatsky, Anant Rai, Anikait Singh, Anthony Brohan, et~al.
\newblock Open x-embodiment: Robotic learning datasets and rt-x models.
\newblock \emph{arXiv preprint arXiv:2310.08864}, 2023.

\bibitem[Iyer et~al.(2024)Iyer, Peng, Dai, Guzey, Haldar, Chintala, and Pinto]{iyer2024open}
Aadhithya Iyer, Zhuoran Peng, Yinlong Dai, Irmak Guzey, Siddhant Haldar, Soumith Chintala, and Lerrel Pinto.
\newblock Open teach: A versatile teleoperation system for robotic manipulation.
\newblock \emph{arXiv preprint arXiv:2403.07870}, 2024.

\bibitem[Khazatsky et~al.(2024)Khazatsky, Pertsch, Nair, Balakrishna, Dasari, Karamcheti, Nasiriany, Srirama, Chen, Ellis, et~al.]{khazatsky2024droid}
Alexander Khazatsky, Karl Pertsch, Suraj Nair, Ashwin Balakrishna, Sudeep Dasari, Siddharth Karamcheti, Soroush Nasiriany, Mohan~Kumar Srirama, Lawrence~Yunliang Chen, Kirsty Ellis, et~al.
\newblock Droid: A large-scale in-the-wild robot manipulation dataset.
\newblock \emph{arXiv preprint arXiv:2403.12945}, 2024.

\bibitem[Fu et~al.(2024)Fu, Zhao, and Finn]{fu2024mobile}
Zipeng Fu, Tony~Z Zhao, and Chelsea Finn.
\newblock Mobile aloha: Learning bimanual mobile manipulation with low-cost whole-body teleoperation.
\newblock \emph{arXiv preprint arXiv:2401.02117}, 2024.

\bibitem[Brohan et~al.(2022)Brohan, Brown, Carbajal, Chebotar, Dabis, Finn, Gopalakrishnan, Hausman, Herzog, Hsu, et~al.]{brohan2022rt}
Anthony Brohan, Noah Brown, Justice Carbajal, Yevgen Chebotar, Joseph Dabis, Chelsea Finn, Keerthana Gopalakrishnan, Karol Hausman, Alex Herzog, Jasmine Hsu, et~al.
\newblock Rt-1: Robotics transformer for real-world control at scale.
\newblock \emph{arXiv preprint arXiv:2212.06817}, 2022.

\bibitem[Fang et~al.(2023)Fang, Fang, Tang, Liu, Wang, Zhu, and Lu]{fang2023rh20t}
Hao-Shu Fang, Hongjie Fang, Zhenyu Tang, Jirong Liu, Junbo Wang, Haoyi Zhu, and Cewu Lu.
\newblock Rh20t: A robotic dataset for learning diverse skills in one-shot.
\newblock \emph{arXiv preprint arXiv:2307.00595}, 2023.

\bibitem[Zhao et~al.(2023)Zhao, Kumar, Levine, and Finn]{zhao2023learning}
Tony~Z Zhao, Vikash Kumar, Sergey Levine, and Chelsea Finn.
\newblock Learning fine-grained bimanual manipulation with low-cost hardware.
\newblock \emph{arXiv preprint arXiv:2304.13705}, 2023.

\bibitem[Sharma et~al.(2018)Sharma, Mohan, Pinto, and Gupta]{sharma2018multiple}
Pratyusha Sharma, Lekha Mohan, Lerrel Pinto, and Abhinav Gupta.
\newblock Multiple interactions made easy (mime): Large scale demonstrations data for imitation.
\newblock In \emph{Conference on robot learning}, pages 906--915. PMLR, 2018.

\bibitem[Bahl et~al.(2023)Bahl, Mendonca, Chen, Jain, and Pathak]{bahl2023affordances}
Shikhar Bahl, Russell Mendonca, Lili Chen, Unnat Jain, and Deepak Pathak.
\newblock Affordances from human videos as a versatile representation for robotics.
\newblock In \emph{Proceedings of the IEEE/CVF Conference on Computer Vision and Pattern Recognition}, pages 13778--13790, 2023.

\bibitem[Jang et~al.(2022)Jang, Irpan, Khansari, Kappler, Ebert, Lynch, Levine, and Finn]{jang2022bc}
Eric Jang, Alex Irpan, Mohi Khansari, Daniel Kappler, Frederik Ebert, Corey Lynch, Sergey Levine, and Chelsea Finn.
\newblock Bc-z: Zero-shot task generalization with robotic imitation learning.
\newblock In \emph{Conference on Robot Learning}, pages 991--1002. PMLR, 2022.

\bibitem[Lynch et~al.(2023)Lynch, Wahid, Tompson, Ding, Betker, Baruch, Armstrong, and Florence]{lynch2023interactive}
Corey Lynch, Ayzaan Wahid, Jonathan Tompson, Tianli Ding, James Betker, Robert Baruch, Travis Armstrong, and Pete Florence.
\newblock Interactive language: Talking to robots in real time.
\newblock \emph{IEEE Robotics and Automation Letters}, 2023.

\bibitem[Mandlekar et~al.(2021)Mandlekar, Xu, Wong, Nasiriany, Wang, Kulkarni, Fei-Fei, Savarese, Zhu, and Mart{\'\i}n-Mart{\'\i}n]{mandlekar2021matters}
Ajay Mandlekar, Danfei Xu, Josiah Wong, Soroush Nasiriany, Chen Wang, Rohun Kulkarni, Li~Fei-Fei, Silvio Savarese, Yuke Zhu, and Roberto Mart{\'\i}n-Mart{\'\i}n.
\newblock What matters in learning from offline human demonstrations for robot manipulation.
\newblock \emph{arXiv preprint arXiv:2108.03298}, 2021.

\bibitem[Wu et~al.(2023)Wu, Escontrela, Hafner, Abbeel, and Goldberg]{wu2023daydreamer}
Philipp Wu, Alejandro Escontrela, Danijar Hafner, Pieter Abbeel, and Ken Goldberg.
\newblock Daydreamer: World models for physical robot learning.
\newblock In \emph{Conference on Robot Learning}, pages 2226--2240. PMLR, 2023.

\bibitem[Mees et~al.(2022)Mees, Hermann, Rosete-Beas, and Burgard]{mees2022calvin}
Oier Mees, Lukas Hermann, Erick Rosete-Beas, and Wolfram Burgard.
\newblock Calvin: A benchmark for language-conditioned policy learning for long-horizon robot manipulation tasks.
\newblock \emph{IEEE Robotics and Automation Letters}, 7\penalty0 (3):\penalty0 7327--7334, 2022.

\bibitem[Mitash et~al.(2023)Mitash, Wang, Lu, Terhuja, Garaas, Polido, and Nambi]{mitash2023armbench}
Chaitanya Mitash, Fan Wang, Shiyang Lu, Vikedo Terhuja, Tyler Garaas, Felipe Polido, and Manikantan Nambi.
\newblock Armbench: An object-centric benchmark dataset for robotic manipulation.
\newblock In \emph{2023 IEEE International Conference on Robotics and Automation (ICRA)}, pages 9132--9139. IEEE, 2023.

\bibitem[Thomas et~al.(2023)Thomas, Cheng, Loynd, Frujeri, Vineet, Jalobeanu, and Kolobov]{thomas2023plex}
Garrett Thomas, Ching-An Cheng, Ricky Loynd, Felipe~Vieira Frujeri, Vibhav Vineet, Mihai Jalobeanu, and Andrey Kolobov.
\newblock Plex: Making the most of the available data for robotic manipulation pretraining.
\newblock In \emph{Conference on Robot Learning}, pages 2624--2641. PMLR, 2023.

\bibitem[Sohn et~al.(2015)Sohn, Lee, and Yan]{sohn2015learning}
Kihyuk Sohn, Honglak Lee, and Xinchen Yan.
\newblock Learning structured output representation using deep conditional generative models.
\newblock \emph{Advances in neural information processing systems}, 28, 2015.

\bibitem[Chi et~al.(2023)Chi, Feng, Du, Xu, Cousineau, Burchfiel, and Song]{chi2023diffusion}
Cheng Chi, Siyuan Feng, Yilun Du, Zhenjia Xu, Eric Cousineau, Benjamin Burchfiel, and Shuran Song.
\newblock Diffusion policy: Visuomotor policy learning via action diffusion.
\newblock \emph{arXiv preprint arXiv:2303.04137}, 2023.

\bibitem[Ebert et~al.(2021)Ebert, Yang, Schmeckpeper, Bucher, Georgakis, Daniilidis, Finn, and Levine]{ebert2021bridge}
Frederik Ebert, Yanlai Yang, Karl Schmeckpeper, Bernadette Bucher, Georgios Georgakis, Kostas Daniilidis, Chelsea Finn, and Sergey Levine.
\newblock Bridge data: Boosting generalization of robotic skills with cross-domain datasets.
\newblock \emph{arXiv preprint arXiv:2109.13396}, 2021.

\bibitem[Walke et~al.(2023)Walke, Black, Zhao, Vuong, Zheng, Hansen-Estruch, He, Myers, Kim, Du, et~al.]{walke2023bridgedata}
Homer~Rich Walke, Kevin Black, Tony~Z Zhao, Quan Vuong, Chongyi Zheng, Philippe Hansen-Estruch, Andre~Wang He, Vivek Myers, Moo~Jin Kim, Max Du, et~al.
\newblock Bridgedata v2: A dataset for robot learning at scale.
\newblock In \emph{Conference on Robot Learning}, pages 1723--1736. PMLR, 2023.

\bibitem[Bharadhwaj et~al.(2023)Bharadhwaj, Vakil, Sharma, Gupta, Tulsiani, and Kumar]{bharadhwaj2023roboagent}
Homanga Bharadhwaj, Jay Vakil, Mohit Sharma, Abhinav Gupta, Shubham Tulsiani, and Vikash Kumar.
\newblock Roboagent: Generalization and efficiency in robot manipulation via semantic augmentations and action chunking.
\newblock \emph{arXiv preprint arXiv:2309.01918}, 2023.

\bibitem[Nair et~al.(2023)Nair, Rajeswaran, Kumar, Finn, and Gupta]{nair2023r3m}
Suraj Nair, Aravind Rajeswaran, Vikash Kumar, Chelsea Finn, and Abhinav Gupta.
\newblock R3m: A universal visual representation for robot manipulation.
\newblock In \emph{Conference on Robot Learning}, pages 892--909. PMLR, 2023.

\bibitem[Vaswani et~al.(2017)Vaswani, Shazeer, Parmar, Uszkoreit, Jones, Gomez, Kaiser, and Polosukhin]{vaswani2017attention}
Ashish Vaswani, Noam Shazeer, Niki Parmar, Jakob Uszkoreit, Llion Jones, Aidan~N Gomez, {\L}ukasz Kaiser, and Illia Polosukhin.
\newblock Attention is all you need.
\newblock \emph{Advances in neural information processing systems}, 30, 2017.

\bibitem[He et~al.(2016)He, Zhang, Ren, and Sun]{he2016deep}
Kaiming He, Xiangyu Zhang, Shaoqing Ren, and Jian Sun.
\newblock Deep residual learning for image recognition.
\newblock In \emph{Proceedings of the IEEE conference on computer vision and pattern recognition}, pages 770--778, 2016.

\bibitem[Perez et~al.(2018)Perez, Strub, De~Vries, Dumoulin, and Courville]{perez2018film}
Ethan Perez, Florian Strub, Harm De~Vries, Vincent Dumoulin, and Aaron Courville.
\newblock Film: Visual reasoning with a general conditioning layer.
\newblock In \emph{Proceedings of the AAAI conference on artificial intelligence}, volume~32, 2018.

\bibitem[Tan and Le(2019)]{tan2019efficientnet}
Mingxing Tan and Quoc Le.
\newblock Efficientnet: Rethinking model scaling for convolutional neural networks.
\newblock In \emph{International conference on machine learning}, pages 6105--6114. PMLR, 2019.

\bibitem[Nair et~al.(2022)Nair, Rajeswaran, Kumar, Finn, and Gupta]{nair2022r3m}
Suraj Nair, Aravind Rajeswaran, Vikash Kumar, Chelsea Finn, and Abhinav Gupta.
\newblock R3m: A universal visual representation for robot manipulation.
\newblock \emph{arXiv preprint arXiv:2203.12601}, 2022.

\bibitem[Radford et~al.(2021)Radford, Kim, Hallacy, Ramesh, Goh, Agarwal, Sastry, Askell, Mishkin, Clark, et~al.]{radford2021learning}
Alec Radford, Jong~Wook Kim, Chris Hallacy, Aditya Ramesh, Gabriel Goh, Sandhini Agarwal, Girish Sastry, Amanda Askell, Pamela Mishkin, Jack Clark, et~al.
\newblock Learning transferable visual models from natural language supervision.
\newblock In \emph{International conference on machine learning}, pages 8748--8763. PMLR, 2021.

\end{thebibliography}

\section*{Checklist}


\begin{enumerate}

\item For all authors...
\begin{enumerate}
  \item Do the main claims made in the abstract and introduction accurately reflect the paper's contributions and scope?
    \answerYes{}
  \item Did you describe the limitations of your work?
    \answerYes{}
  \item Did you discuss any potential negative societal impacts of your work?
    \answerNA{}
  \item Have you read the ethics review guidelines and ensured that your paper conforms to them?
    \answerYes{}
\end{enumerate}

\item If you are including theoretical results...
\begin{enumerate}
  \item Did you state the full set of assumptions of all theoretical results?
    \answerNA{}
	\item Did you include complete proofs of all theoretical results?
    \answerNA{}
\end{enumerate}

\item If you ran experiments (e.g. for benchmarks)...
\begin{enumerate}
  \item Did you include the code, data, and instructions needed to reproduce the main experimental results (either in the supplemental material or as a URL)?
    \answerYes{}
  \item Did you specify all the training details (e.g., data splits, hyperparameters, how they were chosen)?
    \answerYes{}
	\item Did you report error bars (e.g., with respect to the random seed after running experiments multiple times)?
    \answerYes{}
	\item Did you include the total amount of compute and the type of resources used (e.g., type of GPUs, internal cluster, or cloud provider)?
    \answerYes{}
\end{enumerate}

\item If you are using existing assets (e.g., code, data, models) or curating/releasing new assets...
\begin{enumerate}
  \item If your work uses existing assets, did you cite the creators?
    \answerYes{}
  \item Did you mention the license of the assets?
    \answerYes{}
  \item Did you include any new assets either in the supplemental material or as a URL?
    \answerYes{}
  \item Did you discuss whether and how consent was obtained from people whose data you're using/curating?
    \answerYes{}
  \item Did you discuss whether the data you are using/curating contains personally identifiable information or offensive content?
    \answerYes{}
\end{enumerate}

\item If you used crowdsourcing or conducted research with human subjects...
\begin{enumerate}
  \item Did you include the full text of instructions given to participants and screenshots, if applicable?
    \answerNA{}
  \item Did you describe any potential participant risks, with links to Institutional Review Board (IRB) approvals, if applicable?
    \answerNA{}
  \item Did you include the estimated hourly wage paid to participants and the total amount spent on participant compensation?
    \answerNA{}
\end{enumerate}

\end{enumerate}


\clearpage
\appendix
\section{Licensing}
BRMData is released under the open-source MIT license. For more details, please visit \href{https://embodiedrobot.github.io/}{https://embodiedrobot.github.io/}. All personal identity information involved in the dataset has been authorized.

\section{Dataset Accessing}

\begin{itemize}
    \item Link to access BRMData: 
    
    \url{ http://box.jd.com/sharedInfo/1147DC284DDAEE91DC759E209F58DD60}
    \item Password for accessing BRMData: $\texttt{f2ss31}$
\end{itemize}
 
\section{Data Formats}

The dataset format for BRMData is structured to comprehensively capture both visual observations and actions. It includes image data from three cameras, each providing RGB images with a resolution of $480 \times 640 \times 3$. Depth images from the same cameras are also included. Additionally, the dataset records joint positions and efforts for 14 joints, divided equally between the left and right follower arms. This format ensures a detailed representation of the robot's state and actions, facilitating advanced analysis and research in robotic manipulation.

% ...

\begin{center}
\begin{minipage}{0.8\textwidth}
\begin{lstlisting}[basicstyle=\ttfamily\footnotesize, breaklines=true, frame=single, backgroundcolor=\color{lightgray!10}]
[
  {
    "observations": 
    {
      "images":
      {
        "cam_high": (480, 640, 3),
        "cam_left_wrist": (480, 640, 3),
        "cam_right_wrist": (480, 640, 3),
      },
      "images_depth":
      {
        "cam_high": (480, 640),
        "cam_left_wrist": (480, 640),
        "cam_right_wrist": (480, 640),
      },   
      "qpos": (14), # 0-7:left arm, 7-14: right arm
      "qvel": (14), # 0-7:left arm, 7-14: right arm
      "effort": (14) # 0-7:left arm, 7-14: right arm
    },
    "action": (14), # 0-7:left arm, 7-14: right arm
    "base_action": (2) 
  },
  ...
]
\end{lstlisting}
\end{minipage}
\end{center}

\section{Tasks of BRMData}
\label{appendix:tasks}

\subsection{Task Descriptions}
\label{appendix:task_descriptions}

\begin{figure}[h]
  \centering
  \includegraphics[width=0.99\linewidth]{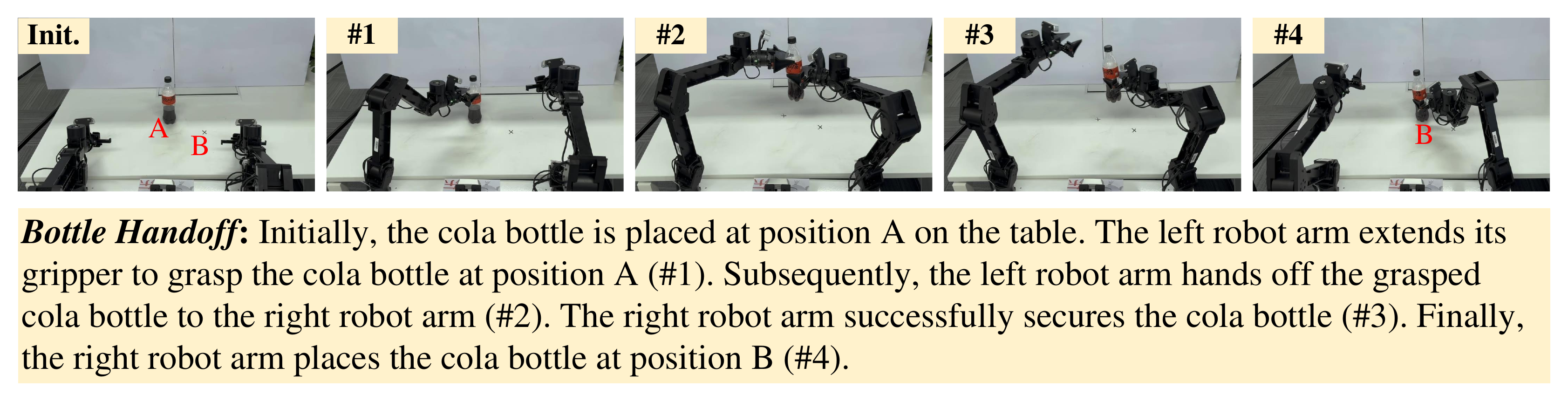}
  \caption{Task Definition of \textbf{\textit{Bottle Handoff}}.}
  \label{bottle_handoff}
\end{figure}

\begin{table}[h]
\renewcommand\arraystretch{1.2}
\centering
\caption{Description of robot manipulation tasks}
\resizebox{0.95\textwidth}{!}{
\begin{tabular}{cp{11cm}}
\toprule
Task              & Task Description                                                                                                         \\ \hline
\textit{Bottle Pick}       & A single robot arm picking up a cola bottle from position A on the table and placing it at position B.                   \\ \hline
\textit{Single Fruit Pick} & Pick up a fruit from a table and place it into a plate.                                                                  \\ \hline
\textit{Plate Place}       & A single robot arm picks up a randomly placed plate from the table and accurately positions it into a fixed dish rack.   \\ \hline
\textit{Fruit Handover}    & A single arm picks up a fruit from a plate and hands it over to a person.                                                \\ \hline
\textit{Bottle Handoff}    & Use two robot arms to transfer a cola bottle.                                                                            \\ \hline
\textit{Cup Place}         & Dual-arm coordination to pick up a cup initially upright on a table, invert it, and place it upside down in a dish rack. \\ \hline
\textit{Multi Fruit Pick}  & Two robot arms collaboratively arrange a plate and sequentially place an orange and an apple into the plate on a table.  \\ \hline
\textit{Wine Wipe} &
  The left arm lifts a wine glass off the table, then the right arm cleans the spilled wine with a cloth. After cleaning, both the wine glass and cloth are placed back on the table. \\ \hline
\textit{Garbage Recycle}   & Pick up a waste item from the table, locate the garbage bin, navigate to it, and deposit the item into the bin.          \\ \hline
\textit{High Five}         & Lift a single arm to high-five an approaching person while moving.                                                       \\ \bottomrule
\end{tabular}
}\label{task_descriptions}
\end{table}

Table \ref{task_descriptions} provides a comprehensive overview of the various robot manipulation tasks included in our dataset. These tasks range from simple object localization, such as picking and placing a bottle (\textit{Bottle Pick}) and single fruit handling (\textit{Single Fruit Pick}), to more complex dual-arm coordination tasks like \textit{Cup Place} and \textit{Multi Fruit Pick}. The tasks are designed to test the robot's ability to perform precise actions, handle objects collaboratively, and interact with humans, such as in the \textit{Fruit Handover} and \textit{High Five} scenarios.

\subsection{Task Definitions}
\label{appendix:task_definitions}

\begin{figure}[h]
  \centering
  \includegraphics[width=0.99\linewidth]{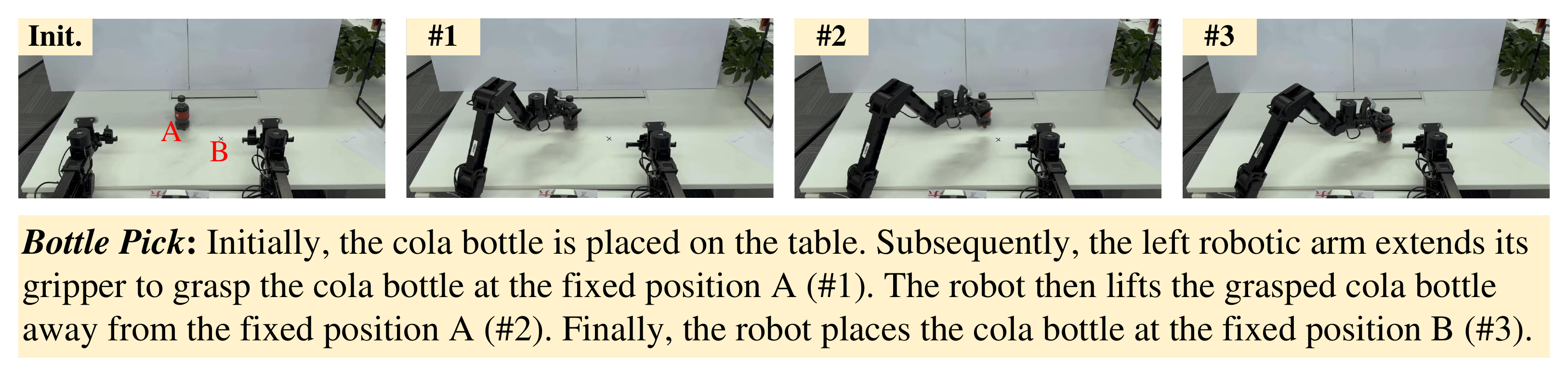}
  \caption{Task Definition of \textbf{\textit{Bottle Pick}}.}
  \label{bottle_pick_task}
\end{figure}

1) As shown in Figure \ref{bottle_pick_task}, the \textbf{\textit{Bottle Pick}} task features a single robot arm executing a precise sequence of movements, transferring a cola bottle from position A to position B on a table on 400 defined steps. This task is designed to assess the arm's capabilities in object gripping and spatial location perception, crucial for automated handling systems. By focusing on the arm's ability to accurately grasp a bottle and then navigate it between two points, the task tests the precision of the arm's motor skills and its sensory feedback mechanisms. The challenge lies in ensuring the arm can consistently and reliably handle objects while adapting to variations in environmental conditions, such as different object orientations and positions, which are key for real-world applications in logistics and manufacturing automation.

2) As shown in Figure \ref{bottle_handoff}, the \textbf{\textit{Bottle Handoff}} task demonstrates a coordinated interaction between two robot arms, transferring a cola bottle across a sequence of 800 steps. This task assesses the collaborative capabilities of bimanual robot manipulation, emphasizing their synchronized movements and precision in object handoffs. The core of this task is the fluid transfer of the bottle from the left to the right robot arm, underscoring the critical need for accurate timing and spatial coordination between the two arms. This manipulation serves as a vital benchmark for the interoperability and collaborative efficiency of bimanual-mobile robots, which are crucial for improving operational harmony and enhancing task execution in complex settings.

\begin{figure}[h]
  \centering
  \includegraphics[width=0.99\linewidth]{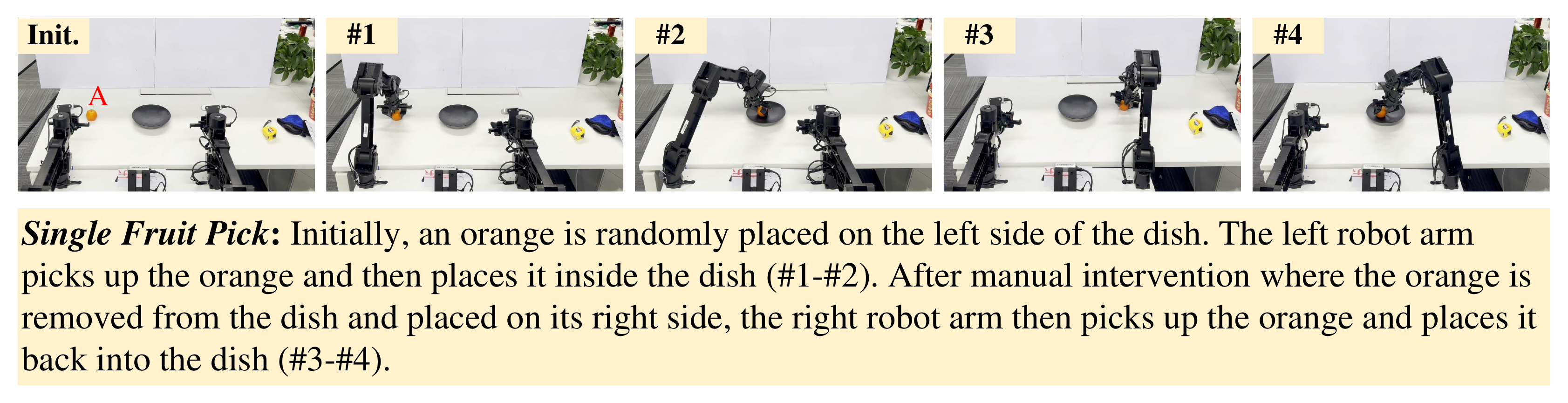}
  \caption{Task Definition of \textbf{\textit{Single Fruit Pick}}.}
  \label{single_fruit_pick}
\end{figure}
3) As shown in Figure \ref{single_fruit_pick}, the \textbf{\textit{Single Fruit Pick}} task simulates human-like decision-making by employing the nearest robot arm to execute the fruit picking manipulation, a process encompassing 400 steps. This task involves an orange initially placed on the one side of a dish, where the nearest robot arm picks it up and places it inside the dish. This approach not only tests each arm's ability to perform precise pick-and-place actions independently, but also emphasizes the efficiency of using spatially advantageous positions, mimicking how humans would typically perform such a task. The task highlights the robot arms' precision, adaptability, and the capability to handle delicate objects within confined spaces, showcasing their potential for applications requiring nuanced and context-aware robotic actions.

\begin{figure}[ht]
  \centering
  \includegraphics[width=0.99\linewidth]{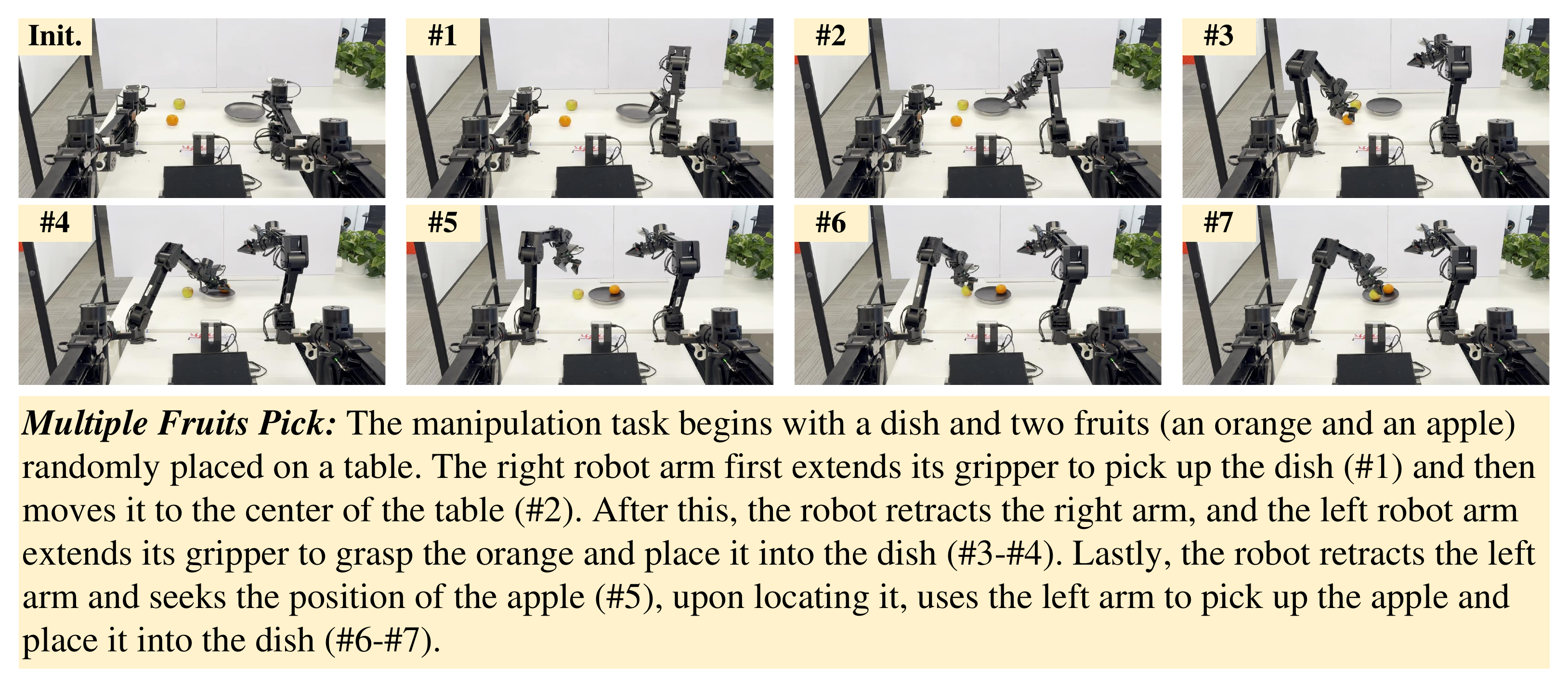}
  \caption{Task Definition of \textbf{\textit{Multi Fruit Pick}}.
}
  \label{multi_fruit_pick}
\end{figure}
4) As shown in Figure \ref{multi_fruit_pick}, the \textbf{\textit{Multi Fruit Pick}} task demonstrates advanced bimanual coordination on a table setup, where a dish and two fruits are initially placed randomly. This 900-step task significantly advances the complexity found in the \textit{Single Fruit Pick} task by requiring sequential manipulations between two robot arms. The right arm initiates by precisely positioning the dish at the center of the table, setting the stage for subsequent manipulations. The left arm then engages, placing the orange followed by the apple into the dish. This sequence not only assesses each arm’s precision in handling pick-and-place tasks, but also underscores the challenge of harmonizing their actions within a shared workspace. The \textit{Multi Fruit Pick} task serves as a balanced evaluation of the robots' bimanual coordination, spatial awareness, and adaptability, vital for complex manipulative tasks in real-world applications.

\begin{figure}[h]
  \centering
  \includegraphics[width=0.99\linewidth]{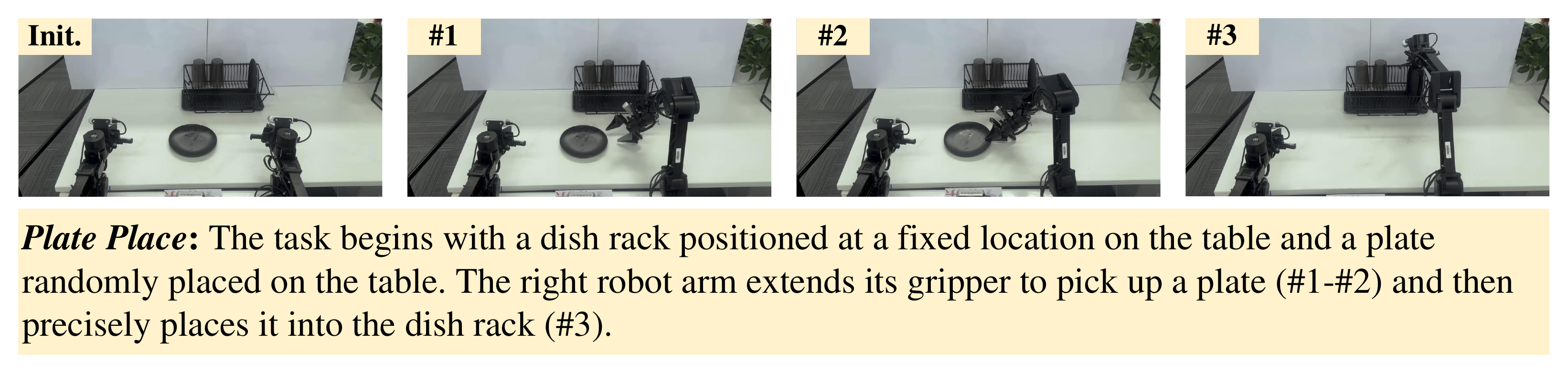}
  \caption{Task Definition of \textbf{\textit{Plate Place}}.
}
  \label{plate_place}
\end{figure}


5) As shown in Figure \ref{plate_place}, the \textbf{\textit{Plate Place}} task comprises 600 action steps, offering a moderate level of complexity that focuses on the robot's ability to handle typical kitchenware with precision. The challenge lies in the robot's ability to detect the random position of the plate, accurately maneuver its arm to grasp it, and then carefully place it on the narrow confines of a draining rack. This tests not only the robot’s perception and localization capabilities but also its end-effector control and fine motor skills. 

\begin{figure}[ht]
  \centering
  \includegraphics[width=0.99\linewidth]{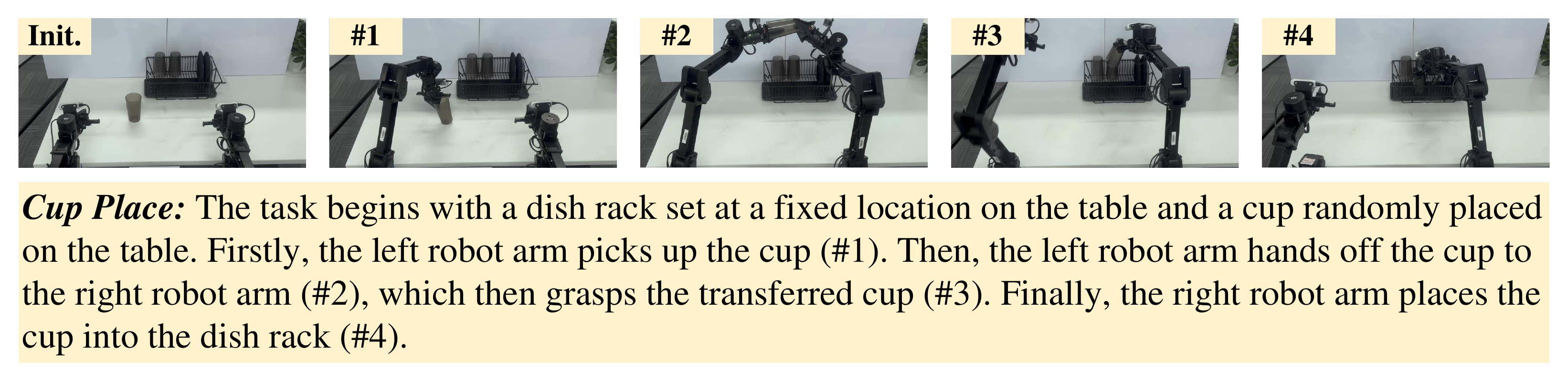}
  \caption{Task Definition of \textbf{\textit{Cup Place}}.
}
  \label{cup_place}
\end{figure}
6) The \textbf{\textit{Cup Place}} task is a sophisticated demonstration of bimanual coordination, where a robot is tasked with placing a randomly positioned cup into a specific slot on a dish rack already containing several cups. This task, which includes 800 action steps, represents a notable escalation in complexity compared to the simpler \textit{Plate Place} task. The key challenge here involves not only the synchronized movement of both arms but also the robot's ability to accurately identify empty slots within the partially filled rack and manipulate the cup to fit snugly upside down in its designated space. This process tests the robot’s cognitive capabilities to perceive its environment and adapt its manipulation strategies accordingly. 


\begin{figure}[ht]
  \centering
  \includegraphics[width=0.99\linewidth]{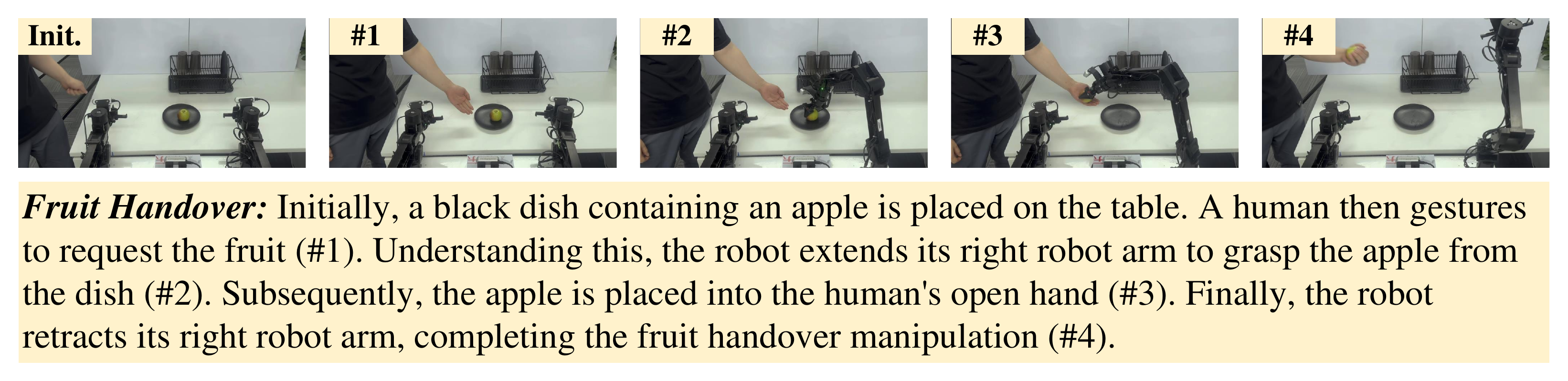}
  \caption{Task Definition of \textbf{\textit{Fruit Handover}}.
}
  \label{fruit_handover}
\end{figure}
7) The \textbf{\textit{Fruit Handover}} task demonstrates a robot interacting directly with a human to transfer fruit, highlighting its potential for practical robotic assistance in real-world scenarios. This task evaluates the collaborative abilities of a single robot arm, focusing on its capacity to understand human gestures and respond by delivering fruit safely and comfortably. The challenge here is twofold: firstly, the robot must accurately detect and interpret the human's hand gesture as a cue to initiate the task. Secondly, the robot needs to execute the task with a level of finesse that ensures the safety and comfort of human interaction, which is critical for applications in settings such as personal care or healthcare where robots assist directly with humans.

\begin{figure}[ht]
  \centering
  \includegraphics[width=0.99\linewidth]{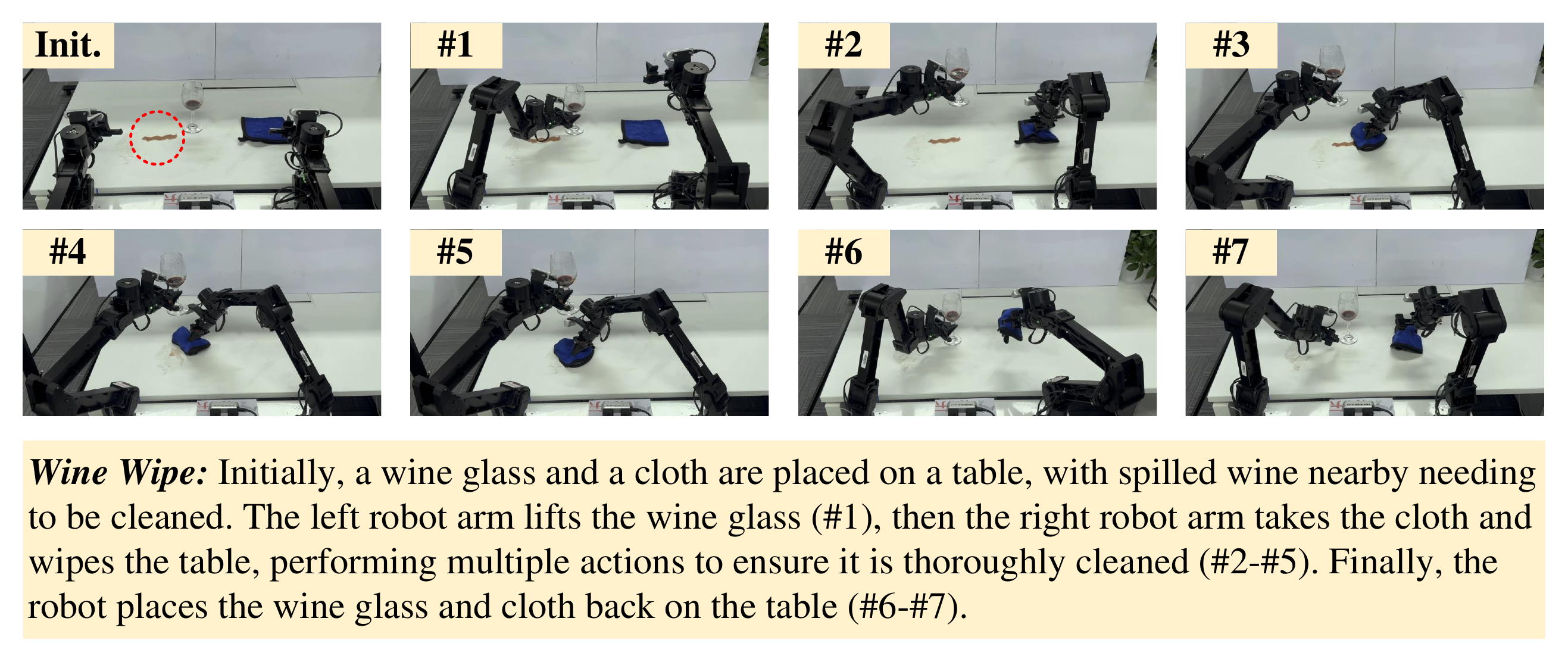}
  \caption{Task Definition of \textbf{\textit{Wine Wipe}}.
}
  \label{wine_wipe}
\end{figure}
8) The \textbf{\textit{Wine Wipe}} task demonstrates a complex bimanual robot manipulation across 1000 action steps, effectively mirroring a typical real-world cleaning scenario. This task is designed to evaluate the robot's finesse and the coordination between its two arms, focusing on their ability to concurrently execute multiple manipulation activities. The task primarily tests the robot's precision and endurance through the right arm's repeated motions required to thoroughly clean the spill, ensuring no residue remains. Additionally, the task challenges the robot's spatial cognition and memory by requiring it to precisely replace the wine glass and cloth to their original positions. This aspect is essential for roles that demand careful handling and reorganization within dynamic and potentially cluttered environments. Thus, the \textit{Wine Wipe} task serves as a crucial benchmark for assessing advanced robot manipulation skills necessary in service-oriented and domestic settings.

\begin{figure}[h]
  \centering
  \includegraphics[width=0.99\linewidth]{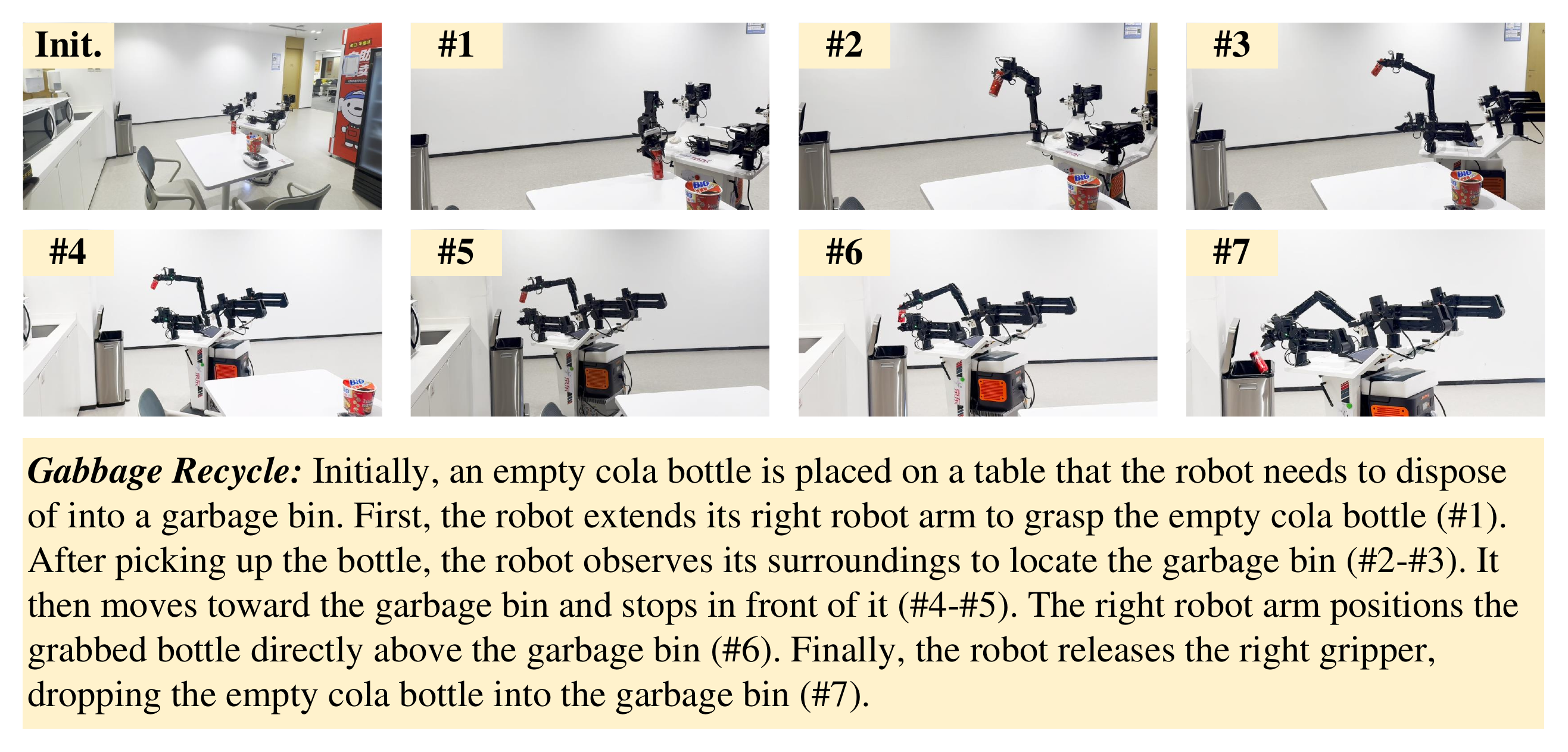}
  \caption{Task Definition of \textbf{\textit{Gabbage Recycle}}.
}
  \label{gabbage_recycle}
\end{figure}
9) The \textbf{\textit{Gabbage Recycle}} task is a critical assessment of the bimanual-mobile robot's ability to integrate mobility with precise manipulative actions, testing not just mechanical skills but also advanced cognitive functions. This task specifically evaluates the robot's proficiency in identifying and interacting with a garbage bin within a spatially complex environment, a significant step beyond static table-top manipulation tasks. The robot must accurately detect the recycle bin’s location amidst other elements, demonstrating effective environment mapping and object recognition. Designed to challenge the robot with a sequence of 1200 steps, this task demands sustained precision over an extended duration, simulating real-world scenarios where robots are expected to perform prolonged tasks.

\begin{figure}[h]
  \centering
  \includegraphics[width=0.99\linewidth]{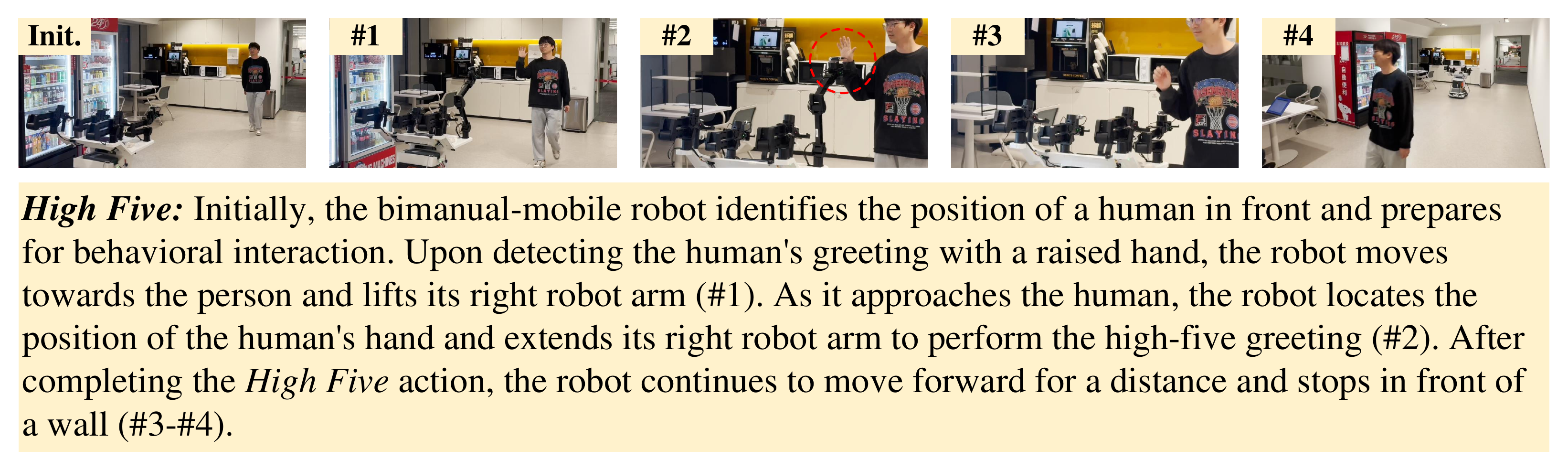}
  \caption{Task Definition of \textbf{\textit{High Five}}.
}
  \label{high_five}
\end{figure}
10) The \textbf{\textit{High Five}} task serves as a critical assessment of the robot's interactive capabilities, particularly in synchronizing its movements with dynamic human actions. This task, involving a robot designed to extend its right arm to high-five a human while simultaneously moving forward, encapsulates 500 action steps. The task is designed to test the robot’s precision in both timing and spatial judgment, a crucial skill set for applications in interactive environments such as homes or offices. By challenging the robot to coordinate its movement with an approaching person, the task not only examines the robot's ability to perform social gestures but also highlights its adaptability and responsiveness to human presence.

\section{Robot Platform}
\label{Robot Platform}
\begin{figure*}[htbp]
    \centering
    \includegraphics[width=1.0\linewidth]{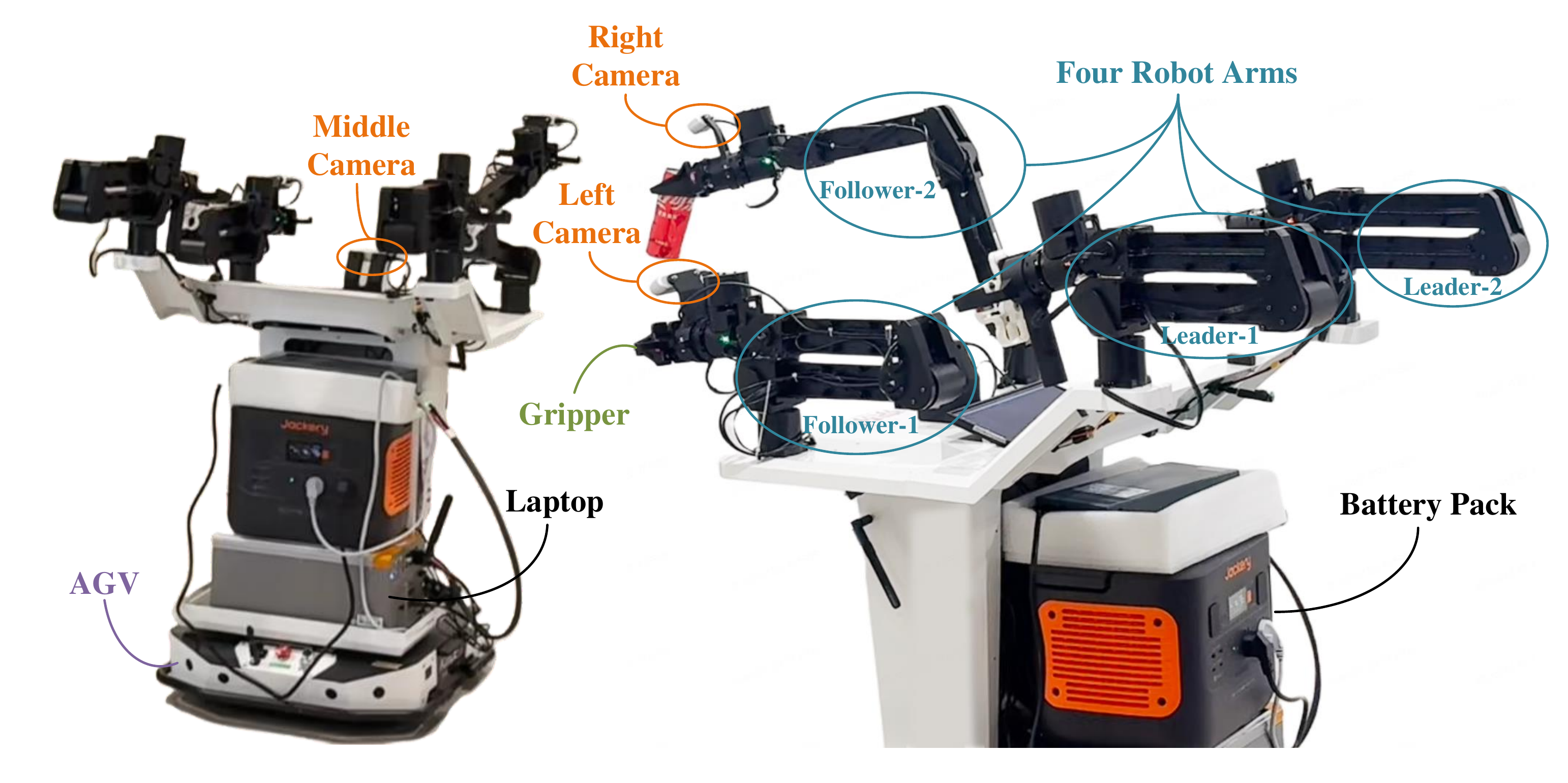}
    \caption{Illustration of robot platform.}
    \label{fig:hardware_architecture}
\end{figure*}

As illustrated in Figure \ref{fig:hardware_architecture}, the data collection hardware platform for BRMData comprises 4 ARX-5 robot arms, each equipped with 7 joints and a parallel gripper.
The \textit{Follower-1} and \textit{Follower-2} robot arms, positioned at the front of the platform, are specifically engineered for advanced bimanual manipulations. Each of both arms features a movable RGBD camera sensor at the wrist. The flexibility afforded by these cameras facilitates variable viewing angles and provides detailed visual feedback, crucial for precision in manipulation tasks. 
Positioned centrally between \textit{Follower-1} and \textit{Follower-2} is a fixed RGBD camera sensor. 
This middle camera provides a consistent and broad visual reference across the manipulation area. It serves as an integral component, enhancing the coordination of the follower arms. By offering a complementary perspective, it enriches the detailed imagery captured by the wrist-mounted cameras.

The \textit{Leader-1} and \textit{Leader-2} robot arms, located at the rear of the platform, are designated for teleoperation by human experts. This configuration facilitates the synchronous replication of human-directed actions from the leader arms to the follower arms, enhancing the platform's capability in executing complex manipulation sequences through a human-teaching data collection strategy.
Supporting the robotic components, the platform includes an Automated Guided Vehicle (AGV) equipped with a two-wheel differential Tracer mobile base, which provides essential mobility for executing mobile manipulation tasks within indoor environment. 
The computational demands of the platform are managed by an industrial computer equipped with an NVIDIA 4090 graphics card, an Intel i7-13700 CPU, and 32GB of RAM. This computing setup ensures efficient processing of the camera data and supports real-time algorithms necessary for effective manipulation and interaction.

\section{Experimental Setup}
\label{appendix:exp_setup}
In the single-task experiments, we perform 20 trials per task to assess the precision and consistency of each method in a controlled environment, with the only variable being the initial positions of objects on the table. This setup allows us to evaluate how well each method handles specific tasks with minimal external interference, where each model is responsible for a single robot manipulation task. 
In the multi-task experiments, we evaluate a single model's ability to manage and solve different robot manipulation tasks individually. Each task undergoes 10 trials to explore how well the model can adapt to and handle various tasks. This approach enables us to assess the model's flexibility and robustness across multiple tasks, reflecting its ability to generalize and perform under varying conditions. 

\section{Robot Manipulation Learning Methods}
\label{appendix:robot_learning_methods}

\subsection{Single-task Robot Manipulation Learning Methods}

\textbf{Action Chunking with Transformers (ACT) \cite{zhao2023learning}:} The ACT method combines a Conditional Variational Auto Encoder (CVAE) \cite{sohn2015learning} with a Transformer architecture \cite{vaswani2017attention} to predict sequences of actions, termed as action chunks, rather than individual actions. This design is particularly suited for complex and precision-demanding tasks. Initially, ACT employs a ResNet-18 \cite{he2016deep} backbone for image processing, maintaining a resolution of $640 \times 480$ pixels for each view.

\textbf{Diffusion Policy (DP) \cite{chi2023diffusion}:} The DP method models robot behavior by characterizing a robot's visuomotor policy through a conditional denoising diffusion process. This approach leverages the stochastic properties of diffusion models to generate diverse and adaptable action sequences.

\subsection{Multi-task Robot Manipulation Learning Methods}

\textbf{MT-ACT \cite{nair2023r3m}:} The Multi-Task ACT (MT-ACT) model augments the ACT framework to proficiently manage multiple tasks simultaneously. In our re-implementation efforts, a FiLM \cite{brohan2022rt,perez2018film} network has been integrated into the ACT model to concurrently execute several manipulation tasks, aligning each task with a corresponding language instruction. These language commands facilitate FiLM-enhanced, language-conditioned learning, enabling the MT-ACT model to dynamically adapt to varying task demands. Detailed descriptions of the specific language commands employed are provided in Table \ref{content}.

\begin{table}[htbp]
\centering
\caption{Language commands for multi-task learning methods}
\begin{tabular}{cm{10cm}}
\toprule
\textbf{Task} & \textbf{Language Command}  \\ \midrule
Gabbage Recycle & grab the cola bottle on the table with the right arm, turn to the trash can, approach the trash can, and throw the cola bottle into the trash can \\  \midrule
High Five & approach the person, raise the right arm, and give he a high five  \\  \midrule
Multi Fruits Pick & grab the plate with the right arm and place it in the center of the table, use the left arm to pick up the orange and put it on the plate, then pick up the peach and put it on the plate  \\ \midrule
Single Fruit Pick & put an orange on the table into the plate  \\  \midrule
Cup Place & pick up the gray cup on the table with the left arm, and then pass the cup to the right arm, place the cup on the shelf with the right arm \\ \midrule
Plate Place & pick up the plate on the table with the right arm and place it on the shelf  \\ \midrule
Fruit Handover & grab the apple from the plate with the right arm and handover it to someone \\ \midrule
Wine Wipe & grab a red wine glass with the left arm and a cloth with the right arm to wipe the scattered red wine on the table  \\ 
\bottomrule
\end{tabular}
\label{content}
\end{table}

\textbf{MT-ACT-EB3:} In this variation of the MT-ACT model, the ResNet-18 image backbone is substituted with an \cite{tan2019efficientnet} model to capitalize on enhanced feature extraction capabilities. Input images are resized to $300 \times 300$ pixels before processing.

\textbf{MT-ACT-R3M:} The MT-ACT-R3M variant employs the R3M model \cite{nair2022r3m}, which is designed for robotic manipulation tasks with pre-trained visual representations. This model adapts the image backbone for better suitability in robotic applications, with input images resized to $224 \times 224$ pixels.

\newpage
\section{Impact of Mobility on Task Complexity}
\label{Impact of Mobility on Task Complexity}
\begin{wraptable}{l}{0.5\textwidth}
\centering
\caption{Performance comparison in stationary versus mobile robot manipulation tasks.}
\begin{tabular}{cccccc}
\toprule
\multirow{2}{*}{Method} & \multicolumn{2}{c}{Wine Wipe}   &    &\multicolumn{2}{c}{Garbage Recycle}\\\arrayrulecolor{gray}\cline{2-3} \cline{5-6} \arrayrulecolor{black} 
& SR  & MES &  & SR & MES  \\ \midrule
ACT & \textbf{100} & \textbf{5.06} & & \textbf{55} & \textbf{2.62} \\
DP  & 80 & 3.38 &  & 40 & 1.57 \\ \bottomrule 
\end{tabular}
\label{table_diff_3}
\end{wraptable}

The experimental results in Table \ref{table_diff_3} demonstrate that tasks requiring mobility are significantly more complex than stationary tasks. In the \textit{Wine Wipe} task, which involves stationary manipulation, ACT achieves a perfect SR of 100\% and an MES of 5.06, while DP achieves an SR of 80\% and an MES of 3.38. However, when the task requires mobility, as in \textit{Garbage Recycle}, performance drops notably: ACT's SR decreases to 55\% and MES to 2.62, and DP's SR decreases to 40\% and MES to 1.57. 
The dual requirement for seamless integration of navigation and manipulation in mobile tasks presents additional challenges, such as maintaining stability while moving and accurately reaching designated locations. These results underscore the challenges in maintaining high performance in tasks that involve mobility, thus demonstrating the robustness and utility of BRMData in evaluating the capabilities of robot learning methods under varied and complex conditions.

\section{Robustness Testing for Robot Manipulation Tasks}
\label{appendix:robustness_test}
\begin{table}[h]
\centering
\renewcommand{\arraystretch}{1.1}
\caption{Robustness test results of single-task models}
\begin{tabular}{cccc}
\toprule
Method & Task              & SR (\%) & MES  \\ \midrule
\multirow{4}{*}{ACT}    & Plate Place       & 40 & 4.29 \\
       & Cup Place         & 60 & 4.57 \\
       & Wine Wipe         & 30 & 1.27 \\
       & Single Fruit Pick & 10 & 0.66 \\ \cline{2-4} 
\multirow{4}{*}{DP}     & Plate Place       & 20 & 1.30 \\
       & Cup Place         & 40 & 2.11 \\
       & Single Fruit Pick & 0  & 0    \\
       & Wine Wipe         & 50 & 2.22 \\ \bottomrule
\end{tabular}
\label{robustness_test}
\end{table}
The repetitive robot manipulation experiments focus on evaluating the model's inference execution and consistency, achieving high SRs due to the low-noise task environment and minimal background changes.
To assess the robustness of robot manipulation learning methods, additional experiments are conducted by introducing significant changes to the tabletop environment and adding distractors. These changes aim to test how well the models maintain performance in the face of unexpected variations and increased complexity. The results, shown in Table \ref{robustness_test}, indicate a significant drop in both SR and MES compared to the initial repetitive tests. For instance, the ACT method’s SR for \textit{Plate Place} drops from 100\% to 40\%, with a corresponding decrease in MES to 4.29. Similarly, DP’s performance in the \textit{Single Fruit Pick} task falls to a 0\% SR, indicating its inability to handle the increased complexity introduced by background changes and distractors. These findings underscore the critical importance of robustness testing in advancing the research in embodied manipulation. High performance in controlled, low-noise environments does not necessarily translate to real-world efficacy. The robustness tests reveal the limitations of the current models, particularly their struggles with environmental variability and unforeseen obstacles. Such testing methodologies are essential to understanding and improving the adaptability and resilience of robot manipulation models. Ultimately, the goal is to develop models that are not only precise and consistent in controlled settings but also resilient and versatile in diverse and challenging real-world environments. This highlights the ongoing need for refinement and enhancement of robot manipulation algorithms to achieve reliable performance in practical applications.

\section{Limitations and Future Work}
\label{appendix:limitations}
 The proposed dataset mainly focuses on household scenes and is not particularly abundant in quantity. In future work, we will collect data from more fields to expand our BRMData, including logistics, medical fields, etc. Moreover, we will always be committed to promoting the development of embodied intelligence, including embodied methods and datasets.

\section{Implementation Details}
\label{appendix:implementation_details}
In this paper, the hyperparameters of the evaluated methods for the proposed dataset are shown in Table \ref{act_params}-\ref{diff_params}, respectively. Meanwhile, in the multi-task robot manipulation learning methods, we adopt the CLIP \cite{radford2021learning} network to encode the language commands. Besides, all the models are trained in a server equipped with 8 NVIDIA L40S graphics cards, an Intel Platinum 8468V CPU, and 1024GB of RAM.

\begin{table}[h]
\renewcommand\arraystretch{1.2}
\centering
\caption{Hyperparameters of ACT}
\begin{tabular}{@{\hspace{30pt}}l@{\hspace{20pt}}l@{\hspace{40pt}}}
\toprule
learning rate      & 4e-5                   \\ 
batch size & 48  \\ 
\# encoder layers       & 4  \\ 
\# decoder layers    & 7 \\ 
feedforward dimension    & 3200 \\ 
hidden dimension         & 512 \\ 
\# heads  & 8  \\ 
chunk size & 32 \\
beta & 10 \\ 
dropout   & 0.1 \\ 
backbone & pretrained ResNet18 \\
\bottomrule
\end{tabular}
\label{act_params}
\end{table}

\begin{table}[h]
\renewcommand\arraystretch{1.2}
\centering
\caption{Hyperparameters of diffusion policy}
\begin{tabular}{@{\hspace{30pt}}l@{\hspace{20pt}}l@{\hspace{40pt}}}
\toprule
learning rate      & 1e-4                   \\ 
batch size & 32  \\ 
chunk size & 64 \\
scheduler & DDIM \\
train and test diffusion steps & 50,10 \\
ema power & 0.75 \\
backbone & pretrained ResNet18 \\
noise predictor & UNet \\
 & RandomCrop (ratio=0.95) \& \\
image augmentation & ColorJitter \& \\
 & RandomRotation \\
\bottomrule
\end{tabular}
\label{diff_params}
\end{table}

\end{document}